\documentclass[11pt,twoside]{toptesi}

\usepackage{graphicx}
\usepackage[pagestyles]{titlesec}
\usepackage[flushleft,justification=centering]{caption}
\usepackage{setspace}
\usepackage[utf8]{inputenc} 
\usepackage[italian]{babel} 
\usepackage[T1]{fontenc} 
\usepackage{blindtext}
\usepackage{graphicx,wrapfig}
\usepackage{graphbox}
\usepackage[export]{adjustbox}
\usepackage{booktabs}
\usepackage[table,xcdraw]{xcolor}
\usepackage{lmodern}
\usepackage{varioref}
\usepackage{url}
\usepackage{array}
\usepackage{paralist}{\obeyspaces\global\let =\space}
\usepackage{verbatim} 
\usepackage[caption=false]{subfig}
\usepackage{tabularx}
\usepackage{amsmath}
\usepackage{amsfonts}
\usepackage{float}
\usepackage{amssymb}
\usepackage{multicol}
\usepackage{multirow}
\usepackage{color}
\usepackage{multirow}
\usepackage{listings}
\usepackage[pass]{geometry}
\usepackage[figuresright]{rotating}
\usepackage{algorithm}
\usepackage{algorithmic}
\usepackage{amsmath}
\usepackage[babel]{csquotes}
\usepackage{hyperref}
\usepackage[backend=bibtex,sorting=none, backref=true]{biblatex}
\usepackage{array}
\usepackage{nameref}
\usepackage{numprint}
\usepackage{microtype}

\newcommand{\setfont}[1]{\fontfamily{iwona}\selectfont \scshape #1}
\DeclareMathOperator{\mean}{E}
\DeclareMathOperator{\variance}{Var}
\DeclareMathOperator{\mse}{MSE}
\DeclareMathOperator{\rmse}{RMSE}

\newcolumntype{P}[1]{>{\centering\arraybackslash}p{#1}}

\makeatletter
\def\thickhline{%
	\noalign{\ifnum0=`}\fi\hrule \@height \thickarrayrulewidth \futurelet
	\reserved@a\@xthickhline}
\def\@xthickhline{\ifx\reserved@a\thickhline
	\vskip\doublerulesep
	\vskip-\thickarrayrulewidth
	\fi
	\ifnum0=`{\fi}}
\makeatother

\newlength{\thickarrayrulewidth}
\hypersetup{%
    pdfpagemode={UseOutlines},
    bookmarksopen,
    pdfstartview={FitH},
    colorlinks,
    linkcolor={black}, 
    citecolor={black}, 
    urlcolor={black} 
}

\makeatletter
\newcommand\listofcodes{%
 \iffrontmatter\else\frontmattertrue\fi
 \if@openright\cleardoublepage\else\clearpage\fi
 \begingroup\def\chapter##1{\@schapter}
 \phantomsection 
 \lstlistoflistings 
 \endgroup
} 
\makeatother

\bibliography{bibliography} 

\begin{document}
	
\newpage
\pagestyle{empty} 
\noindent

\begin{figure}\doublespacing
	\mbox{
				\begin{minipage}{.20\textwidth}
					\includegraphics[height=3.3cm]{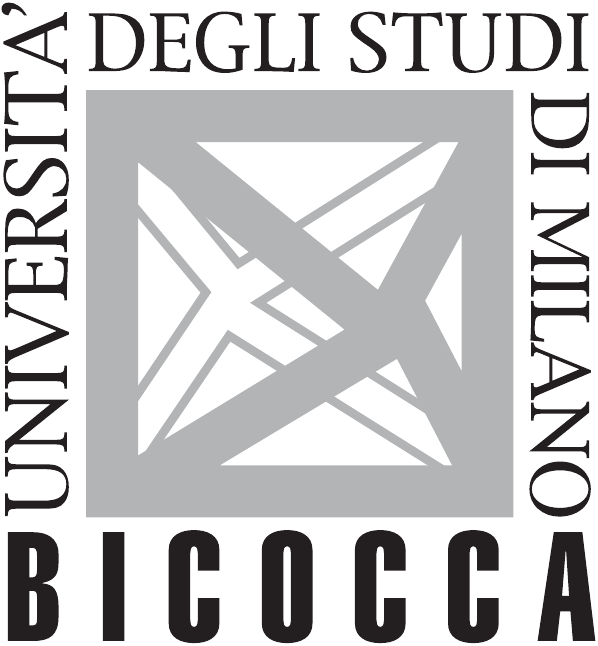}
				\end{minipage}%
				\quad\quad
				\begin{minipage}[c]{.90\textwidth}
					{Università degli Studi di Milano-Bicocca}\\
					{  \textbf{Scuola di Scienze}}\\
					{\textbf{Dipartimento di Informatica, Sistemistica e Comunicazione}}\\
					{  \textbf{Corso di Laurea in Informatica}}
				\end{minipage}%
	}

\end{figure}

\begin{center}
	\vspace{35mm}
\doublespacing\textbf{\huge RETI NEURALI PER~L’APPRENDIMENTO DEI TRATTI DELLA PERSONALITÀ DAL LINGUAGGIO NATURALE }\\
	
\end{center}

\vspace{30mm}
\onehalfspacing 

\begin{tabular}{ll}
	\textbf{Relatore: } & {Prof. Stella Fabio Antonio}\\
	\textbf{Co-relatore: } & {Dott. Marelli Marco}
\end{tabular}

\vspace{5mm}

\begin{flushright}\onehalfspacing 
	\textbf{Relazione della prova finale di:}\\
	{Giorgia Adorni}\\
	{Matricola 806787}\end{flushright}

\vspace{25mm}
\begin{center} \textbf{Anno Accademico 2017-2018 }\end{center}



\frontmatter

\vspace*{\stretch{1}}
\begin{flushright}
\noindent
\textit{A mio padre, per il suo sostegno quotidiano.}\\
\textit{Ad Elia, per tutto il supporto e l'amore dimostrato.} \\
\end{flushright}
\vspace*{\stretch{6}}
\cleardoublepage

%


\ringraziamenti
Grazie al mio relatore Fabio Stella, per avermi trasmesso la sua passione e per avermi fornito gli strumenti necessari per intraprendere questo percorso.\\ 
Grazie ai ragazzi del Laboratorio MAD (Models and Algorithms for Data \& Text Mining), per tutti i loro consigli.


\sommario

La personalità è considerata come uno degli argomenti di ricerca più influenti in psicologia poiché predittiva di molti esiti consequenziali come la salute mentale e fisica, ed è in grado di spiegare il comportamento umano.
Grazie alla diffusione dei Social Network come mezzo di comunicazione, sta diventando sempre più importante sviluppare modelli che possano leggere automaticamente e con precisione l'essenza di individui basandosi esclusivamente sulla scrittura. 
\\
In particolare, la convergenza tra scienze sociali e informatiche ha portato i ricercatori a sviluppare approcci automatici per estrarre e studiare le informazioni ''nascoste'' nei dati testuali presenti in rete.
La natura di questo progetto di tesi è altamente sperimentale, e la motivazione alla base di questo lavoro è presentare delle analisi dettagliate sull'argomento, in quanto allo stato attuale non esistono importanti indagini che si basino interamente su testo in linguaggio naturale.
\\
L'obiettivo è identificare un adeguato spazio semantico che permetta di definire sia la personalità dell'oggetto a cui un determinato testo si riferisce, sia quella dell'autore. Punto di partenza è un dizionario di aggettivi che la letteratura psicologica definisce come \emph{marker} dei cinque grandi tratti di personalità, i Big Five.
\\
In questo lavoro siamo partiti dall'implementazione di reti neurali  fully-connected come base per capire come modelli semplici di Deep Learning possano fornire informazioni sulle caratteristiche nascoste della personalità. 
\\
Infine, utilizziamo una classe di algoritmi distribuzionali inventati nel 2013 da \emph{Tomas Mikolov}, che consistono nell'utilizzo di una rete neurale convoluzionale in grado di imparare, in modo non supervisionato, i contesti delle parole.
In questo modo costruiamo un embedding in cui sono contenute le informazioni semantiche del testo, ottenendo una sorta di “geometria del significato” in cui i concetti sono tradotti in relazioni lineari.
Con quest'ultimo esperimento ipotizziamo che uno stile di scrittura individuale sia in gran parte accoppiato con i tratti della sua personalità.


\tableofcontents

\listoffigures

\listoftables


\introduzione
La personalità è un fattore chiave che influenza le interazioni, i comportamenti e le emozioni delle persone. Al giorno d'oggi, essa viene considera come uno degli argomenti di ricerca più influenti in psicologia.

La crescente immersione negli ambienti digitali e la diffusione dei social network come mezzo di comunicazione, ha contribuito alla creazione di un enorme quantità di dati, anche chiamati \emph{Big Data}, aprendo la necessità allo sviluppo di modelli automatici in grado di leggere con precisione l'essenza degli individui basandosi esclusivamente sulla scrittura.

L'esigenza di produrre analisi sempre più velocemente ha imposto lo sviluppo di metodi meccanici per selezionare e interpretare i dati, favorendo la ricerca nel campo dell'apprendimento automatico o \emph{Machine Learning} \cite{samuel1959some}.

Nello specifico, il \emph{Data Mining} è un approccio che consiste nell'individuazione d’informazioni significative tramite l'applicazione di algoritmi in grado di determinare le associazioni ``nascoste'' tra di esse  \cite{chakrabarti2006data,franklin2005elements}. 

Una sua forma particolare è il \emph{Text Mining}, nell'ambito del quale si sono sviluppate metodologie che consentono ai computer di confrontarsi con il linguaggio umano, di elaborarlo e comprenderlo \cite{tan1999text}.

L'interesse nello studio delle informazioni digitali e le abilità necessarie per farlo non sempre coincidono tra gli scienziati sociali. Di conseguenza, tale ricerca viene generalmente affidata a scienziati e ingegneri informatici, facilitando la scoperta di modelli che non sarebbe possibile individuare ed offrendo l'opportunità di instaurare collaborazioni interdisciplinari.
\\

La maggior parte degli attuali studi automatici di rilevamento della personalità si sono concentrati sulla teoria dei \emph{Big Five} come quadro per studiare le caratteristiche intrinseche dell'essere umano \cite{barrick1991big}.
Secondo questo modello esistono cinque dimensioni fondamentali dei tratti, stabili nel tempo e condivisi a livello interculturale. Le cinque caratteristiche, note appunto come i ``Grandi Cinque'', sono Openness (apertura all'esperienza), Conscientiousness (coscienziosità), Extraversion (estroversione), Agreeableness (gradevolezza), Neuroticism (nevroticismo), riconosciuti dall'acronimo OCEAN.

Sviluppare un modello accurato e aprire questa domanda di ricerca avrebbe implicazioni significative in diversi ambiti della sociologia, ma non solo.

\section*{Struttura della tesi}
Di seguito si passano in rassegna gli argomenti affrontati capitolo per capitolo.

\begin{itemize}
	\item [Nel capitolo \setfont{\nameref{chap:contesto}}] vengono introdotti i concetti teorici alla base del lavoro, in particolare viene introdotta la teoria dei Big Five. 
	\item [Nel capitolo \setfont{ \nameref{chap:RetiNeurali}}] si descrivono le principali tecniche di Deep Learning, in particolare soffermandosi sulle architetture adottate negli esperimenti. 
	\item [Nel capitolo \setfont{\nameref{chap:formulazione}}] viene definito il problema ed illustrati approcci e strumenti risolutivi.
	\item [Nel capitolo \setfont{\nameref{chap:esperimenti}}] viene presentata una panoramica degli esperimenti effettuati e una relativa analisi dei risultati.
	\item [Nel capitolo \setfont{\nameref{chap:conclusioni}}] vengono esposte le considerazioni finali.
\end{itemize}

\section*{Motivazioni}
\label{sec:motivazione}
\addcontentsline{toc}{section}{Motivazione}

La natura di questo progetto di tesi è altamente sperimentale. Le motivazioni che hanno portato alla realizzazione di questo lavoro sono la presentazione di analisi dettagliate sull'argomento, in quanto allo stato attuale non esistono importanti indagini di questo tipo.

\section*{Contributi}
\label{sec:contributi}
\addcontentsline{toc}{section}{Contributi}

I dati che verranno utilizzati per definire lo spazio semantico e testare la sua funzionalità sono messi a disposizione da Yelp Dataset Challenge, che contiene \numprint{5200000} recensioni relative a \numprint{174000} attività commerciali di 11 aree metropolitane nel mondo.


\mainmatter
\chapter{Contesto}
\label{chap:contesto}

Con il termine \emph{personalità} si intende l'insieme delle caratteristiche psichiche e dei comportamentali abituali --- inclinazioni, interessi e passioni --- che definiscono e differenziano ogni individuo, nei vari contesti ed ambienti in cui la condotta umana si sviluppa \cite{corr2009cambridge,sadock2000comprehensive}.

La tradizione di studi psicologici relativi alla personalità è una delle più rilevanti della psicologia contemporanea, un campo in cui si susseguono studi empirici, teorici e storici, volti a comprendere la natura dell'identità personale nel contesto biologico e sociale di sviluppo.
Essa tenta di spiegare le tendenze che sono alla base delle differenze comportamentali, ed ogni gruppo di pensiero tenta di concettualizzare la personalità entro modelli diversi --- adoperando metodi, approcci, obiettivi e modalità d'analisi --- anche molto dissonanti fra loro.

Una significativa parte della psicologia che studia le differenze individuali, analizza e valuta la personalità attraverso specifici test volti ad individuarne i tratti.

\section{Big Five}
\label{sec:big5}

Le teorie della personalità basate sui tratti definiscono la personalità come l'insieme delle caratteristiche che stabiliscono il comportamento di una persona. 

La teoria dei Grandi Cinque (o Big Five) risulta essere uno dei modelli più condivisi e testati, sia a livello teorico che empirico.
McCrae e Costa identificano cinque grandi dimensioni in cui può essere suddivisa la personalità \cite{goldberg1993structure,costa2008revised}:
\begin{itemize}
	\item L'\emph{apertura all'esperienza} o ``openness'' (creativo/curioso vs. coerente/cauto) è intesa come attitudine alla ricerca di stimoli culturali e di pensiero esterni al proprio contesto ordinario.
	Essa riflette il grado di curiosità intellettuale, la creatività o una preferenza per la novità; può inoltre essere percepita come imprevedibilità o mancanza di concentrazione. \\ 
	Individui con un'elevata apertura perseguono l'auto-realizzazione, cercando esperienze intense ed euforiche. Viceversa, coloro che hanno una bassa apertura cercano di ottenere soddisfazione attraverso la perseveranza.
	
	\item La \emph{coscienziosità} o ``conscientiousness'' (organizzato vs. negligente) è una tendenza caratterizzata dall'organizzazione, precisione e affidabilità. Un soggetto contraddistinto da questa attitudine, preferisce un comportamento pianificato piuttosto che spontaneo.\\  
	Spesso l'alta coscienziosità viene percepita come testardaggine e ossessione, mentre la bassa coscienziosità è associata alla flessibilità e alla spontaneità, ma può anche apparire come mancanza di affidabilità.
		
	\item L'\emph{estroversione} o ``extraversion''  (estroverso/energetico vs. solitario/riservato) è intesa come grado di entusiasmo negli atteggiamenti che si adottano e tendenza a cercare la stimolazione in compagnia degli altri.\\ 
	L'alta estroversione è spesso percepita come una ricerca di attenzioni e prepotenza. La bassa estroversione causa una personalità riservata, riflessiva, che può essere avvertita come distaccata.  
	
	\item La \emph{gradevolezza} o ``agreeableness'' (amichevole/compassionevole vs. provocatorio/distaccato) è indicata come quantità e qualità delle relazioni interpersonali che la persona intraprende, orientate al prendersi cura dell'altro. È una tendenza ad essere compassionevoli e collaborativi piuttosto che sospettosi e antagonisti. \\
	L'alta gradevolezza è spesso vista come ingenuità o sottomissione. Le persone con scarsa gradevolezza sono spesso competitive o sfidanti, e possono essere intese come inaffidabili.

	\item Il \emph{nevroticismo} o ``neuroticism'' (sensibile/nervoso vs. sicuro/fiducioso), è una misura di resistenza a stress di tipo psicologico, come l'ansietà e l'irritabilità, ma si riferisce anche al grado di solidità emotiva e di controllo degli impulsi.\\
	Un'alta stabilità si manifesta in una personalità calma che però può essere vista come poco interessante e indifferente. Una bassa stabilità esprime reattività e dinamicità in individui che spesso possono essere percepiti come instabili o insicuri. 
\end{itemize}
Queste dimensioni sono state individuate a partire da studi psico-lessicali, secondo cui le cinque dimensioni corrisponderebbero alle macro-categorie più usate nel linguaggio per descrivere le diversità fra individui.\\
Le regioni cerebrali che codificano i vari tratti di personalità sono spesso collegate alle regioni responsabili della comunicazione verbale e scritta. 

Un grande numero di prove di ricerca hanno supportato il modello a cinque fattori, che sembra essere condiviso a livello interculturale --- Cina, Giappone, Italia, Ungheria, Turchia \cite{triandis2002cultural}.

Le dimensioni di Big Five predicono accuratamente il comportamento e vengono utilizzate sempre più spesso per aiutare i ricercatori a comprendere l'estensione dei disturbi psicologici come ansia e depressione \cite{saulsman2004five}.

Uno dei vantaggi principali di questo approccio è che consente di concentrare l'attenzione solo sulle dimensioni di base piuttosto che studiare centinaia di tratti.

\chapter{Reti Neurali}
\label{chap:RetiNeurali}

Nel campo dell'apprendimento automatico, o \emph{machine learning}, una rete neurale artificiale  in inglese \emph{Artificial Neural Network} (ANN),  è un modello matematico basato sulla semplificazione delle reti neurali biologiche \cite{samuel1959some}.

Una rete neurale può essere considerata come un sistema dinamico avente la topologia di un grafo orientato, i cui nodi modellano i neuroni in un cervello biologico, mentre gli archi rappresentano le sinapsi (interconnessioni di informazioni).

Ogni connessione può trasmettere un segnale da un neurone artificiale a un altro, i quali sono tipicamente aggregati in strati. Gli stimoli vengono ricevuti da un livello di nodi d'ingresso, detto unità di elaborazione, che elabora il segnale e lo trasmette ad altri neuroni ad esso collegati.

\section{Modello}
\label{sec:modello}

Le reti neurali possono essere viste come semplici modelli matematici che definiscono una funzione $f:X\rightarrow Y$. 

La funzione di rete di un neurone $f(x)$ è definita come una composizione di altre funzioni $g_i(x)$, che possono a loro volta essere scomposte in altre funzioni.

Una rappresentazione ampiamente utilizzata per la descrizione di ANN tradizionali è la \emph{somma ponderata}, mostrata nell'equazione \ref{eq:modellomat}.

\begin{equation}
f(x)=K \bigg( \sum_{i}w_ix_i +b\bigg)
\label{eq:modellomat}
\end{equation}

Ogni segnale in ingresso $x_i$ viene moltiplicato ad un corrispondente peso $w_i$, che assume valore positivo o negativo a seconda che si voglia eccitare o inibire il neurone.  
Il bias $b$ varia secondo la propensione del neurone ad attivarsi, influenzandone l'uscita.
Inoltre, viene applicata una funzione predefinita $K$, detta anche \emph{funzione di attivazione}, illustrata nella seguente sezione.

\begin{figure}[htb]
	\centering
	{\includegraphics[width=.7\textwidth]{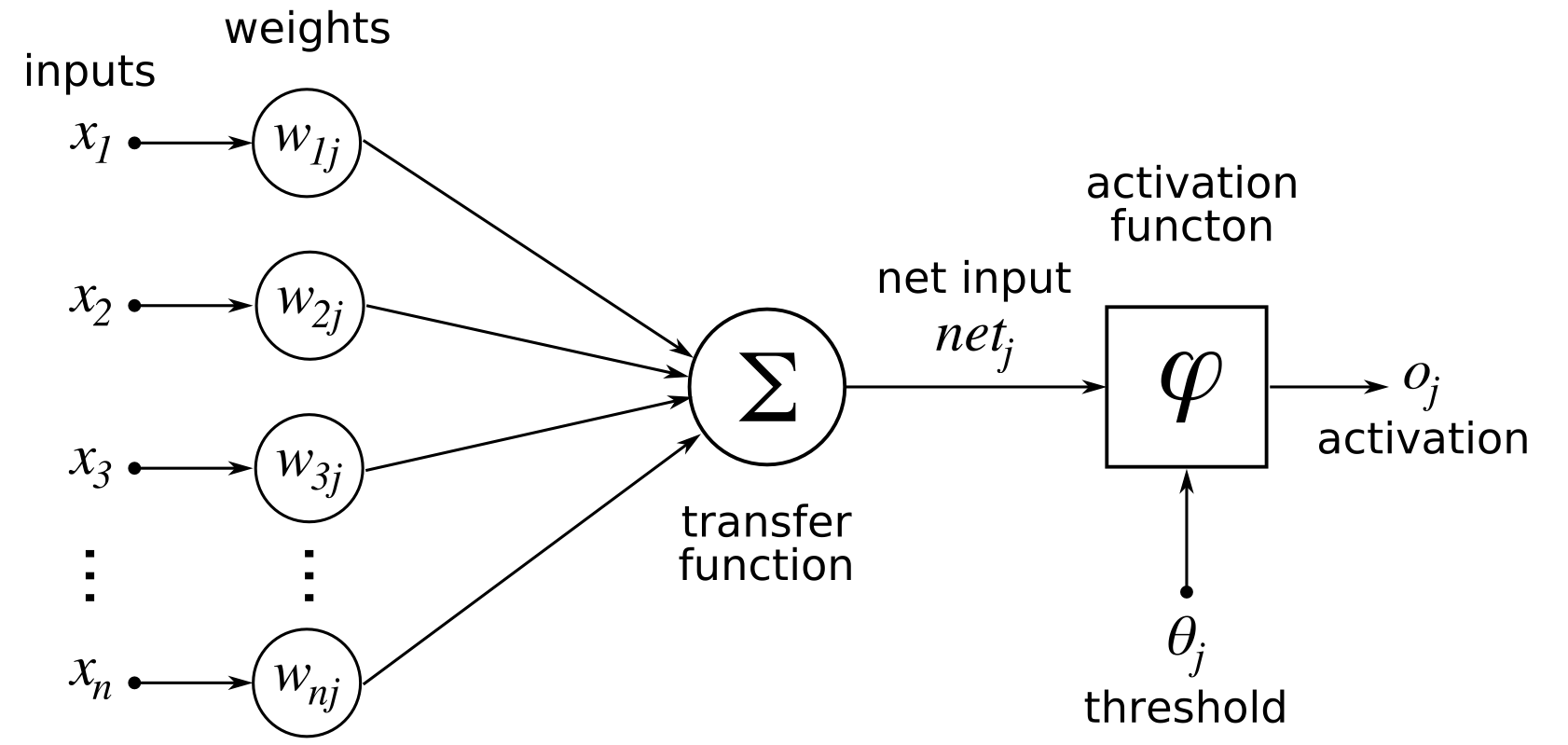}} 
	\caption{Artificial Neuron Model}
	\label{fig:Modello matematico di un neurone artificiale}
\end{figure}

\subsection{Funzioni di attivazione}
\label{subsec:fattivazione}
Una funzione di attivazione è una componente fondamentale del modello. Essa consente alla rete di imparare trasformazioni non lineari, in modo da essere in grado di calcolare problemi non banali utilizzando un limitato numero di nodi.

Una delle funzioni più utilizzate è la \emph{sigmoide} $\sigma(x)$, la quale modella la frequenza degli stimoli emessi, da neurone inattivo, $\sigma(x)=0$, a neurone completamente saturo con una frequenza di attivazione massima, $\sigma(x)=1$.

\begin{equation}
\sigma(x) = \frac{1}{1+e^{-x}}
\label{eq:sigmoid}
\end{equation}

Negli ultimi anni è diventata molto popolare la \emph{Rectified Linear Unit} (ReLU) \cite{nair2010rectified,hahnloser2000digital,hahnloser2003permitted,glorot2011deep}, definita dalla seguente equazione:
\begin{equation}
f (x) = \max(0, x)= \begin{cases}
x \quad \mbox{se } x>0\\
0 \quad \mbox{altrimenti}
\end{cases}
\label{eq:relu}
\end{equation}
Questa funzione azzera tutti i valori negativi, mentre ritorna invariati quelli positivi.

Essa viene utilizzata per la sua capacità di accelerare notevolmente il processo di ottimizzazione, inoltre la sua implementazione risulta semplice ed efficiente.

\begin{figure}[htb]
	\centering
	\subfloat[][\emph{Sigmoide}]
	{\includegraphics[width=.45\textwidth]{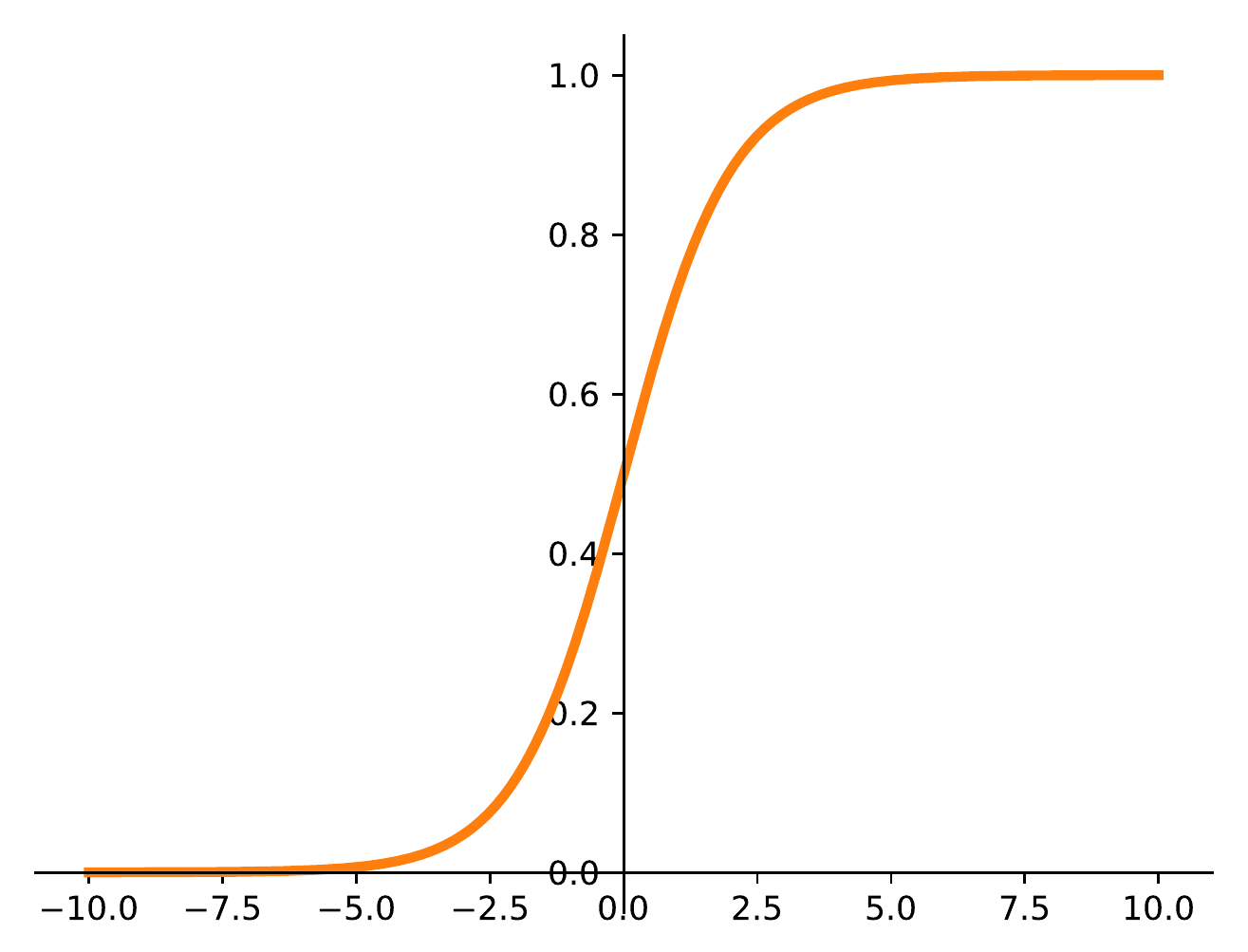}}
	\quad
	\subfloat[][\emph{ReLU}]
	{\includegraphics[width=.45\textwidth]{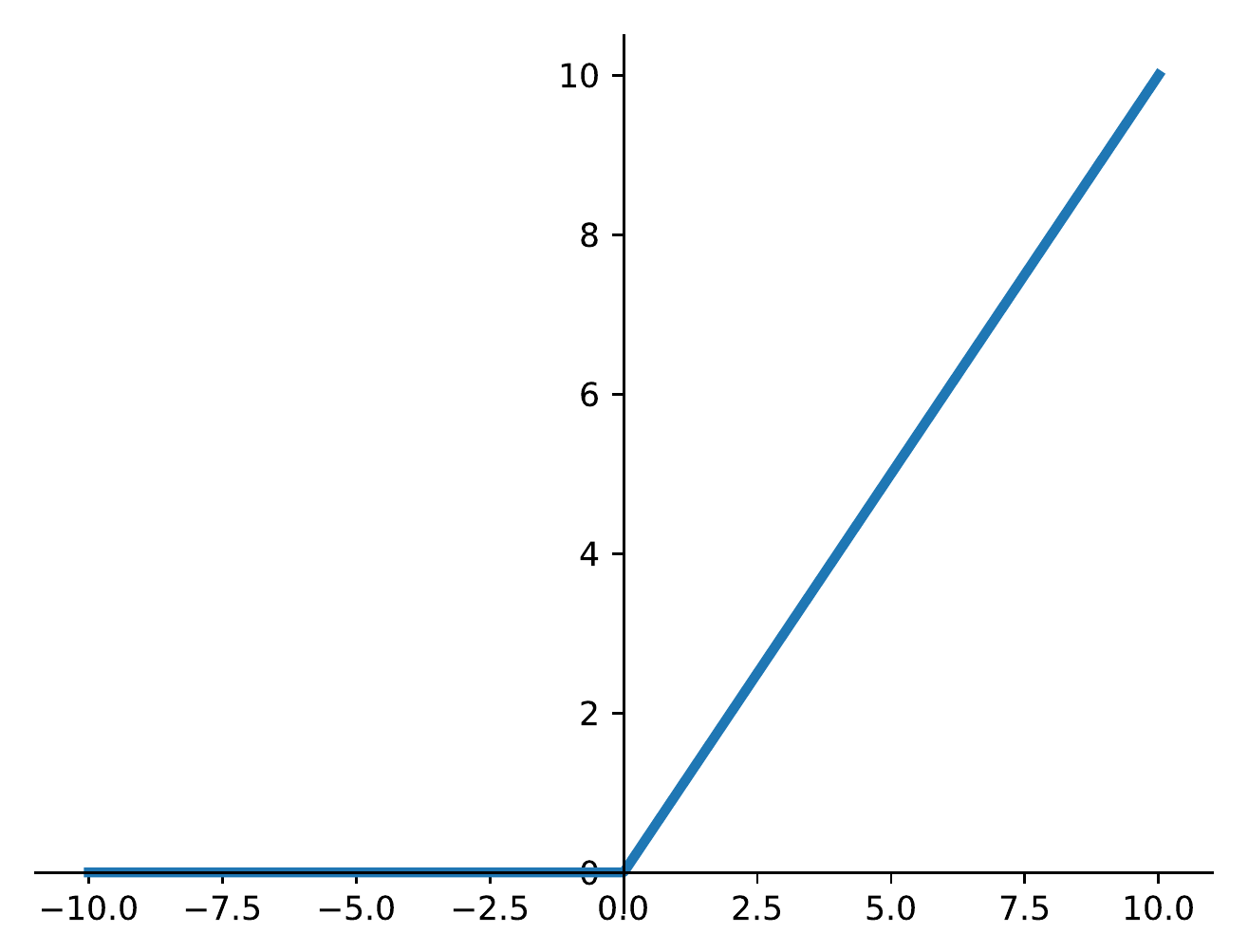}} 
	
	\caption{Andamento di due funzioni di attivazione}
	\label{fig:subfig}
\end{figure}

\section{Architetture}
\label{sec:architetture}

I neuroni vengono organizzati in una struttura detta architettura della rete.
I dati, partendo da un livello iniziale, chiamato layer di input, attraversano i multipli strati interni della rete, gli hidden layer, raggiungendo l'ultimo livello detto layer di output.

Quando i collegamenti tra i neuroni formano una struttura senza cicli si parla di reti \emph{feed-forward} \cite{svozil1997introduction}.

\subsection{Layer fully-connected}
\label{subsec:fc}

Un’architettura molto comune nelle reti neurali è una struttura ``densa'', che utilizza \emph{layer fully-connected}, in cui tutti i neuroni del livello precedente sono collegati ad ogni neurone dello strato successivo \cite{sainath2015convolutional}.

Lo scopo di un layer completamente connesso è imparare combinazioni non lineari di feature ad alto livello provenienti dal layer precedente. 
Una struttura di questo tipo è però caratterizzata da un numero di connessioni che cresce molto velocemente, causando un accrescimento del numero di parametri che la rete deve apprendere.
Questo comporta un aumento del costo computazionale e un alto rischio di overfitting, approfondito nella Sezione~\ref{subsec:overfitting}.

Per questo motivo questi vengono spesso sostituiti dai layer convoluzionali.

\subsection{Layer convoluzionali}
\label{subsec:cnn}
Una rete neurale convoluzionale, in inglese \emph{Convolutional Neural Network} (CNN), è una classe di reti artificiali avanzate e \emph{feed-forward}, composte da uno o più strati convoluzionali seguiti da una serie di livelli completamente connessi \cite{kim2014convolutional}.

Una convoluzione può essere considerata come una funzione a finestra scorrevole, detta \emph{kernel} o filtro, applicata a una matrice. 
Ogni livello applica filtri diversi, in genere centinaia o migliaia, e combina i loro risultati. 
Durante la fase di addestramento, una CNN impara automaticamente i valori dei suoi filtri in base all'attività che si desidera eseguire.

Due parametri che definiscono il comportamento di un layer convoluzionale sono lo \emph{stride}, che rappresenta il passo di 
convoluzione e viene utilizzato per ridurre le dimensioni spaziali dell'output e il \emph{padding}, che definisce il comportamento dei neuroni lungo 
i bordi dei dati di input. Se viene scelto un padding \emph{valid}, l'output contiene solo i neuroni la cui regione di convoluzione è completamente contenuta nei dati; nel caso di padding \emph{same}, invece, l'output mantiene la stessa dimensione dell'input, utilizzando zero come valore per i dati mancanti.

Le reti convoluzionali sono adatte per elaborare dati visivi e altri dati bidimensionali, ed hanno mostrato ottimi risultati nel riconoscimento di immagini e nella manipolazione del linguaggio naturale \cite{manning1999foundations}.
In quest'ultimo caso, l'input è costituito da frasi o documenti rappresentati come una matrice, in cui ogni riga corrisponde a un token, in genere una parola. Tipicamente, questi vettori sono dei word embeddings (rappresentazioni a bassa dimensione) come word2vec \cite{mikolov2013distributed}, ma potrebbero anche essere vettori unici che indicizzano la parola in un vocabolario. 

Nel \emph{Natural Language Processing} (NLP) utilizziamo filtri che scorrono su righe complete della matrice (parole), pertanto, la loro ``larghezza'' è solitamente uguale alla quella della matrice di input. L'altezza, o la dimensione della regione, può variare, ma le finestre tipicamente scorrono su 2-5 parole per volta. 

Risulta che le CNN applicate ai problemi di NLP funzionino abbastanza bene, un esempio è il modello \emph{bag-of-words} che è stato l'approccio standard per anni e ha portato a risultati piuttosto buoni \cite{wallach2006topic}.

Le CNN sono più facili da addestrare e hanno molti meno parametri da stimare. 
Il minor numero di connessioni e pesi di questa architettura, rende gli strati convoluzionali relativamente economici in termini di memoria e potenza di calcolo necessari.

\subsection{Layer di pooling}
\label{subsec:maxpool}

Il \emph{pooling} è un processo alquanto comune nelle reti neurali, la cui funzione è ridurre progressivamente la dimensionalità spaziale per diminuire la quantità di parametri e la complessità computazionale della rete, mantenendo le informazioni più salienti e controllando anche l'overfitting.

Il layer di pooling opera indipendentemente su ogni slice di profondità dell'input e lo ridimensiona spazialmente, usando l'operazione MAX sul risultato di ogni filtro \cite{karpathy2016cs231n}. 

Una proprietà del pool è che fornisce una matrice di output a dimensione fissa --- in genere richiesta per la classificazione. Ciò consente di utilizzare frasi di dimensioni variabili e filtri di dimensioni variabili, ma ottenere sempre le stesse dimensioni di output da inserire in un classificatore.

\subsection{Batch Normalization}
\label{subsec:normalization}

Durante la fase di addestramento del modello, i parametri di ogni substrato vengono ottimizzati al fine di minimizzare l'errore finale.
Ad ogni iterazione, in ciascuno strato avviene una variazione dell'output, corrispondente ad una variazione nei valori in ingresso al livello successivo.
Questo può rappresentare un problema per la rete, che deve adattare i propri strati ad un continuo cambiamento nell'input. 

Per aumentare la stabilità della rete neurale, velocizzare il training e migliorarne le performance, generalmente viene applicata ad ogni strato una \emph{Batch Normalization}  \cite{ioffe2015batch}, la quale normalizza l'uscita di un precedente livello sottraendo il valore medio di un batch e dividendo il risultato per la sua deviazione standard.
\begin{equation}
	\hat{x}^{(k)}=\frac{x^{(k)}-\mean[x^{(k)}]}{\sqrt{\variance[x^{(k)}]}}
\end{equation}

L'applicazione di questa operazione ad ogni input potrebbe cambiare ciò che il layer può rappresentare. Per questo motivo viene assicurato che la trasformazione inserita nella rete possa rappresentare l'identità, in modo da poter annullare il potenziale effetto della Batch Normalization, nel caso in cui fosse l'azione ottimale.
Dunque viene prevista l'aggiunta di due parametri --- ``deviazione standard'' $\gamma$ e ``media'' $\beta$ --- che la rete impara assieme ai parametri del modello originale, come mostrato nell'equazione \ref{eq:batchnorm}
\begin{equation}
	y^{(k)}=\gamma^{(k)}\hat{x}^{(k)}+\beta^{(k)}\mbox{.}
	\label{eq:batchnorm}
\end{equation}

La \emph{Batch Normalization} può essere utilizzata sia su reti \emph{feed-forward}, sia sulle \emph{reti convoluzionali}.
In questo secondo caso, media e varianza vengono calcolate per ogni filtro.

\section{Apprendimento}
\label{sec:apprendimento}
Per insegnare alla rete a risolvere un determinato problema, occorre una fase di addestramento in cui vengono condotte una serie di osservazioni per stabilire quali valori assegnare ad ogni parametro della rete e trovare un modello ottimale.

Questo processo di apprendimento viene strutturato come un problema di ottimizzazione in cui lo scopo è minimizzare una \emph{funzione di costo}, che misura la distanza tra una soluzione particolare ed una ottima. 

\subsection{Funzione di costo}
\label{subsec:loss}

Una funzione di costo mappa un evento ad un numero reale, il quale ne rappresenta intuitivamente il ``costo''.

Nella strategia adottata si utilizzeranno due diverse funzioni obiettivo: l'\emph{errore quadratico medio} per risolvere il problema di regressione, e la \emph{Softmax Cross Entropy} per il compito di classificazione. Mentre per la costruzione dell'embedding verrà applicata la \emph{Noise Contrastive estimation}.

\subsubsection{Mean Squared Error}
\label{subsubsec:MSE}

Per il task il cui compito è prevedere dei valori reali è comune calcolare lo scostamento tra la quantità prevista dalla rete $(\hat{Y})$ e i valori osservati $Y$ (ground truth). 
 
L'\emph{errore quadratico medio} (MSE) di uno stimatore misura la media dei quadrati degli errori, e viene calcolato come

\begin{equation}
\mse = \frac{1}{n}\sum_{i=1}^{n}(Y_i-\hat{Y}_i)^2 \mbox{.}
\end{equation}

con $n$ numero delle previsioni \cite{wang2009mean}.

\subsubsection{Softmax Cross Entropy}
\label{subsubsec:sce}

Softmax è una funzione di loss comunemente utilizzata per la classificazione. In particolare viene applicata allo strato finale della rete ed addestrata in un regime di entropia incrociata \cite{tang2013deep}.\\
L'entropia incrociata è un indicatore che può essere utilizzato per misurare l'accuratezza delle previsioni. 

Date due variabili casuali discrete $p$ e $q$ definiamo l'entropia nel modo seguente:
\begin{equation}
	H(p,q)=-\sum_{x} p(x) \log q(x)
\end{equation}

in cui $p_{x}$ è la ``vera'' probabilità o distribuzione, mentre $q_{x}$ è la distribuzione ``innaturale'' ottenuta a partire dal modello corrente.

Nella pratica la \emph{Cross Entropy} viene calcolata empiricamente ipotizzando l'equiprobabilità degli eventi, poiché $p$ è ignota. 
Di conseguenza, è ridefinibile come segue

\begin{equation}
H(q)=-\frac{1}{N}\sum_{x} \log q(x)
\end{equation}

dove N è il numero di eventi osservati.

L'obiettivo principale di questa funzione è rendere il risultato della Softmax campionata uguale a quella vera. L'algoritmo si concentra sulla selezione di campioni specifici dalla distribuzione data per ottenere la loss desiderata \cite{liu2016large}.  
L'utilizzo della funzione Softmax ha un effetto considerevole sulle prestazioni. 

\subsubsection{Noise Contrastive estimation}
\label{subsubsec:nce}

La stima contrastiva del rumore è una strategia utilizzata nell'ambito della modellazione linguistica o per la generazione di word embedding dati in input dei corpus molto ampi.

La funzione obiettivo del modello \emph{skip-gram} cerca di trovare rappresentazioni di parole che siano utili per predire le parole circostanti, meglio chiamati contesti, in una frase o in un documento. 
Data una sequenza di parole di addestramento, la funzione obiettivo massimizza la probabilità media di log
\begin{equation}
\frac{1}{T}\sum_{t=1}^{T} \sum_{-c \leq j \leq c, j \neq 0} \log p(w_{t+j} | w_T)
\end{equation}

dove $c$ è la dimensione del contesto di training \cite{mikolov2013distributed}. 

Nell'implementazione del modello \texttt{word2vec}, la formulazione standard dello \emph{skip-gram} definisce la precedente probabilità di log ricorrendo alla funzione Softmax:
\begin{equation}
p_{\theta}(w_{O} | w_{I}) = \frac{\exp({v^{\prime}_{w_{O}}}^{\top} v_{w_{I}})}{\sum_{w=1}^{W} \exp({v^{\prime}_{w}}^{\top} v_{w_I})}
\end{equation}

dove $v_w$ e $v'_w$ sono le rappresentazione vettoriali di input e output e $W$ è il numero delle parole del vocabolario \cite{dyer2014notes}.

In questo modo la previsione di una data parola a partire da un contesto risulta essere un compito computazionalmente intenso, poiché vi sono operazione che coinvolgono l'intero dizionario.

Di conseguenza, un alternativa alla funzione Softmax è l'applicazione della \emph{Noise Contrastive estimation} con campionamento negativo \cite{liu2016classification, viswesvaran2000measurement}. Essa consente un allenamento più veloce e rappresentazioni vettoriali migliori per le parole frequenti.

\subsection{Algoritmi di ottimizzazione}
\label{subsec:optimizer}

Gli algoritmi di ottimizzazione sono necessari per minimizzare il risultato di una determinata funzione obiettivo, la quale dipende dai parametri che il modello deve imparare durante l'addestramento. 

Vengono utilizzate varie strategie e algoritmi di ottimizzazione per aggiornare e calcolare i valori appropriati e ottimali di tale modello, i quali influenzano fortemente l'efficacia del processo di apprendimento.

L'entità dell'aggiornamento è determinata dal tasso di apprendimento $\eta$, in inglese \emph{learning rate}, che garantisce la convergenza al minimo globale, per superfici di errore convesse, e ad un minimo locale, per superfici non convesse. 

\subsubsection{Stochastic Gradient Descent}
\label{subsubsec:SGD}

La discesa del gradiente, in inglese \emph{Gradient Descent} (GD), è un algoritmo iterativo per l'ottimizzazione di funzioni \cite{ruder2016overview}.

Viene utilizzato principalmente per eseguire gli aggiornamenti dei pesi in un modello di rete neurale nel seguente modo
\begin{equation}
\theta = \theta - \eta \nabla J(\theta)
\end{equation}
dove $\eta$ rappresenta il tasso di apprendimento, $\nabla J(\theta)$ è il gradiente della funzione di loss $J(\theta)$ rispetto al parametro $\theta$. 

La tradizionale discesa gradiente, o \emph{Batch Gradient Descent} (GD), calcola il gradiente dell'intero set di dati, eseguendo un solo aggiornamento. Di conseguenza il processo di addestramento può risultare lento e difficile da controllare per i set di dati che sono molto grandi. 

I problemi che si verificano con questo algoritmo vengono risolti applicando una sua variante: la {discesa stocastica del gradiente}, in inglese \emph{Stochastic Gradient Descent} (SGD).
Questa tecnica esegue un aggiornamento alla volta dei parametri per ognuno degli esempi di training
\begin{equation}
\theta = \theta - \eta \nabla J(\theta; x(i); y(i))
\end{equation}

dove $x(i)$ e $y(i)$ sono le coppie di esempi usati per l'addestramento.

Questa tecnica risulta essere molto più veloce di quella classica. 
A causa dei frequenti aggiornamenti, i parametri presentano un'alta varianza, inoltre la funzione loss oscilla tra diverse intensità. Questo favorisce la scoperta di nuovi minimi locali, complicando però la convergenza all'ottimo globale.

\subsubsection{Adagrad}
\label{subsubsec:adagrad}

\emph{Adagrad} è un metodo di apprendimento adattativo che aggiusta il learning rate sulla base dei parametri \cite{duchi2011adaptive} .
In questo algoritmo, la dimensione degli aggiornamenti è grande per parametri associati a caratteristiche poco ricorrenti e piccola per quelli più frequenti. Per questo motivo viene considerato adatto alla gestione di dati sparsi.

Adagrad modifica il tasso di apprendimento generale $\eta$ ad ogni istante di tempo $t$ per ogni parametro $\theta(i)$, sulla base dei gradienti che sono stati calcolati per $\theta(i)$.

Il vantaggio principale di questo metodo è che non è necessario regolare manualmente la frequenza di apprendimento e nella maggior parte delle implementazioni viene usato un valore predefinito --- per esempio \numprint{0,001} --- e lasciato invariato.\\
La principale debolezza consiste nell'accumulo dei ``gradienti quadrati'' nel denominatore, finché ogni termine aggiunto è positivo \cite{ruder2016overview}. Di conseguenza il tasso di apprendimento si riduce fino al punto in cui l'algoritmo non è più in grado di acquisire ulteriori conoscenze. 

\subsection{Paradigmi di apprendimento}
\label{subsec:Paradigmi di apprendimento}

Gli algoritmi di apprendimento sono principalmente suddivisi in due categorie:
\begin{itemize}
	\item[\bfseries supervisionato] --- alla rete viene presentato un training set preparato da un ``insegnante esterno'', composto da coppie significative di valori (input, output atteso).
	
	Quando alla rete neurale viene fornito l'input dall'ambiente, l'insegnante calcola l'output desiderato corrispondente, addestrando la rete mediante un algoritmo (tipicamente quello di back propagation \cite{horikawa1992fuzzy}). 
	
	La rete impara a riconoscere la relazione incognita che lega le variabili di ingresso e uscita, in modo da prevedere il valore di output per qualsiasi valore di ingresso, basandosi solo su una casistica di corrispondenze (coppie input-output).
	
	\item[\bfseries non supervisionato] --- alla rete vengono presentati solo i valori di input, mentre non sono messe a disposizione le informazioni di ritorno dell'ambiente sui valori obiettivo che si vogliono ottenere in risposta o riguardo la correttezza dell'output fornito.
	
	La rete è in grado di individuare da sola pattern, caratteristiche, similarità e regolarità statistiche nei dati di input, acquisendo la capacità di dividerli in cluster rappresentativi che sviluppino delle rappresentazioni interne, senza usare confronti con output noti.
	
	In questo caso, gli algoritmi che modificano i pesi della rete fanno riferimento solo ai dati contenuti nelle variabili di ingresso.
	
	Questo è un tipo di apprendimento autonomo senza controllo esterno sull'errore. È un approccio adatto per ottimizzare le risorse nel caso in cui non si conoscano a priori i gruppi in cui dividere l'input.
\end{itemize}

\section{Train, Validation e Test Sets}
\label{sec:set}
Per misurare le prestazioni di una rete neurale dopo la fase di apprendimento, viene creato un test set formato da coppie non utilizzate per il training e validation set.\\
Vengono generalmente definiti: 
\begin{itemize}
	\item Training set --- sul quale viene eseguito l'algoritmo di apprendimento.
	\item Validation set --- viene utilizzato per regolare i parametri, selezionare le features e prendere decisioni per quanto riguarda l'algoritmo di apprendimento.
	\item Test set --- si utilizza per valutare le performance dell'algoritmo, ma non per prendere decisioni su quale algoritmo di apprendimento o parametri utilizzare. 
\end{itemize}

Una volta definiti i set, ci si concentrerà sul miglioramento delle prestazioni del training e validation set. 

Generalmente la dimensione del test set è un terzo di quella del training. Esso è composto da input critici su cui la risposta della rete deve essere buona. 
Questo funziona bene quando sono messi a disposizione un numero limitato di esempi, ma nell'era dei Big Data, dove i problemi di apprendimento automatico consistono di più di un miliardo di campioni, la frazione di dati allocati agli insiemi di sviluppo e test è ridotta, nonostante il valore assoluto di esempi sia maggiore.

Vengono utilizzate diverse tecniche statistiche per valutare la bontà di un modello.

\subsection{Evaluation metric}
\label{subsec:EvaluationMetric}

Le metriche di valutazione misurano le prestazioni di un modello, discriminando la bontà dei risultati ottenuti.
Vengono presi in considerazione diversi tipi di metriche per valutare i modelli. La scelta della metrica dipende completamente dal tipo di modello e dal piano di implementazione. 

I modelli predittivi si distinguo in due principali categorie: si parla di \emph{regressione} quando l'output da prevedere è continuo, o di \emph{classificazione} nel caso in cui l'output sia nominale o binario. 
\subsubsection{Regressione}
\label{subsubsec:regressione}

L'{scarto quadratico medio}, in inglese \emph{Root Mean Squared Error} (RMSE), è la metrica di valutazione più popolare utilizzata nei problemi di regressione. Questo parametro aiuta a fornire risultati affidabili, mostrando correttamente la grandezza del termine di errore.
La metrica RMSE è definita dalla seguente equazione
\begin{equation}
	\rmse = \sqrt{\frac{\sum_{i=1}^{N}({Y}_i - \hat{Y}_i)^2}{N}} \mbox{.}
	\label{eq:rmse}
\end{equation}

con $\hat{Y}$ quantità prevista dalla rete, $Y$ i valori osservati e $N$ numero delle previsioni.
\subsubsection{Classificazione}
\label{subsubsec:classificazione}

Nei problemi di classificazione, in particolare quella binaria, gli output sono 0 o 1. \\
Una \emph{matrice di confusione}, nota anche come matrice di errore, è una tabella $2 \times 2$, generalizzabile ad una $N \times N$ per problemi ad $N$ classi, che consente la visualizzazione delle prestazioni di un algoritmo di apprendimento supervisionato. 

Ogni colonna della matrice rappresenta le istanze previste di una classe mentre ciascuna riga quelle osservate. 

\begin{figure}[t]
	\centering
	{\includegraphics[width=.3\textwidth]{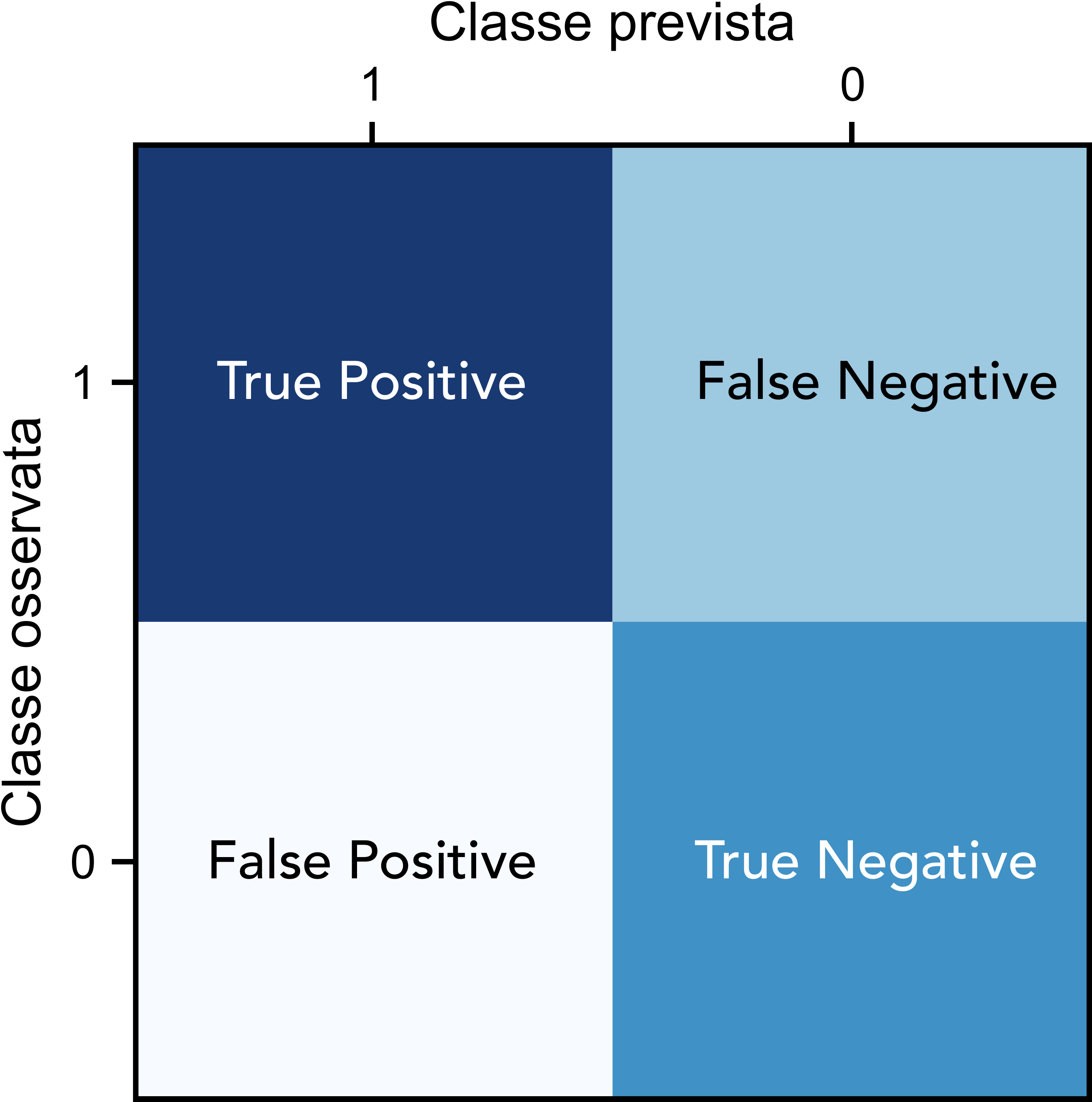}}
	\caption{Visualizzazione di una matrice di confusione}
	\label{fig:confmat}
\end{figure}

L'individuazione di falsi positivi (casi negativi identificati come positivi), falsi negativi (casi positivi identificati come negativi), veri positivi (casi positivi correttamente identificati) e veri negativi (casi negativi correttamente identificati), mostrati nella Figura~\ref{fig:confmat}, consentono  un'analisi più dettagliata della semplice proporzione di classificazioni corrette.

È possibile estrarre da questa tabella le seguenti misure di performance:
\begin{itemize}
	\item l'\emph{accuracy}, o accuratezza, del modello, che consiste nella porzione rispetto al totale delle previsioni corrette
	\begin{equation}
		\frac{\mbox{True Positive} + \mbox{True Negative}}{\mbox{True Positive} + \mbox{True Negative} + \mbox{False Positive} + \mbox{False Negative}}
	\end{equation} 
	\item la \emph{precision}, o precisione, cioè la porzione dei casi positivi identificati correttamente
		\begin{equation}
	\frac{\mbox{True Positive}}{\mbox{True Positive} + \mbox{False Positive}}
	\end{equation}
	
	\item la \emph{recall}, ovvero la porzione dei casi positivi reali correttamente identificati
		\begin{equation}
	\frac{\mbox{True Positive}}{\mbox{True Positive} + \mbox{False Negative}}
	\end{equation}
\end{itemize}

\subsection{Overfitting}
\label{subsec:overfitting}

L'\emph{overfitting} è ``la produzione di un'analisi che corrisponde esattamente a un particolare insieme di dati o ne presenta forti similarità; può determinare l'impossibilità di adattamento a nuovi dati o compromettere l'affidabilità delle predizioni sulle osservazioni future''.

La rete neurale deve avere la capacità di comprensione del modello statistico dei dati. In presenza di overfitting essa memorizza i dati del training set e non è quindi in grado di generalizzare su nuovi dati. L'essenza di questo problema consiste nell'estrarre inconsapevolmente parte della variazione residua (cioè il rumore) come se quella variazione rappresentasse la sottostante struttura del modello \cite{burnham2003model}.

Per ridurre la possibilità di overfitting esistono diverse tecniche, come la convalida incrociata (Cross Validation), la regolarizzazione o l'\emph{early stopping}, che consiste nell'utilizzo di un validation set di coppie non usate nel training set per la misurazione dell'errore.

\chapter{Formulazione}
\label{chap:formulazione}

\section{Definizione del problema }
\label{sec:problem}

Partendo da materiale testuale presente in rete, in particolare un dataset messo a disposizione da \emph{Yelp Dataset Challenge} contenente \numprint{5200000} reviews relative a \numprint{174000} business di 11 aree metropolitane nel mondo, l'obiettivo è quello di estrarre da questi dati caratteristiche di personalità.\\ 
Il nostro scopo è di identificare un adeguato spazio semantico che permetta di definire la personalità dell'oggetto target a cui un determinato testo si riferisce.

\section{Descrizione del dataset}
\label{sec:dataset}

Come punto di partenza, viene messo a disposizione un dizionario di 637 aggettivi, che la letteratura psicologica definisce come marker dei cinque grandi tratti di personalità noti come \emph{Big Five}.
In particolare, questo vocabolario associa ad ogni aggettivo un vettore di cinque elementi in cui ogni elemento corrisponde al grado di presenza o assenza di una determinata caratteristica.
\begin{figure}[H]
	\centering
\begin{tabular}{lccccc}
	\toprule
	 \textbf{Adjective} \quad & \multicolumn{5}{c}{\textbf{OCEAN}} \\
	
\midrule
	Active  & 0,053194 & 0,237406 & 0,365915 & 0,116700 & -0,058669  \\
	Angry  & -0,004604 & -0,038453 & 0,020755 & -0,294754 & 0,590114 \\
	Boring & -0,069877 & -0,099754 & -0,478821 & -0,236462 & 0,118821\\
	\rule{7pt}{0\normalbaselineskip} \dots &   \dots 		&			 \dots &			\dots &			 \dots & \dots \\
	\bottomrule
\end{tabular}
\captionof{table}{Esempio di dizionario OCEAN}
\label{tab:ocean}
\end{figure}

\section{Calcolo della Ground Truth}
\label{sec:GroundTruth}

Per poter addestrare correttamente il modello è necessario avere una Ground Truth, ovvero una label associata ad ogni input della rete. Senza questa informazione sarebbe impossibile ottimizzare il modello e valutarne la validità. 

Si procede eliminando dal dataset tutte le sentences non contenenti almeno uno degli aggettivi presenti nel dizionario OCEAN. In seguito verranno utilizzati i dati contenuti all'interno del vocabolario per creare una mappatura diretta tra ogni frase e il corrispondente vettore di personalità, in cui ogni elemento sarà calcolato come la media del valore di ogni aggettivo presente nel testo.

\section{Preprocessing}
\label{sec:preprocessing}
Gli algoritmi di apprendimento automatico non sono in grado funzionare direttamente con il testo non elaborato, è quindi necessario eseguire in primis un preprocessamento del database di testi e in seguito convertire i dati in numeri, nello specifico, in vettori di numeri.\\
Alla fine di questo processo il dataset ottenuto sarà suddiviso in tre corpora: 
\begin{itemize}
	\item $70\%$ training set : contenente \numprint{4351900} reviews utilizzate dalla rete per l'apprendimento;
	\item $10\%$ validation set: contenente \numprint{621700} reviews;
	\item $20\%$ testing set: contenente \numprint{1243000} reviews.
\end{itemize}
La divisione dei tre dataset dovrà mantenere una buona distribuzione fra le diverse classi.

\subsection{Natural Language Processing}
\label{subsec:nlp}
Il \emph{Natural Language Processing} (NLP) è un insieme di tecniche di computer science e linguistica che ricorrono a dei calcolatori per analizzare il linguaggio umano.
\\

Dal punto di vista sintattico, al dataset viene applicata la seguente serie di operazioni:
\begin{itemize}
	\item \textbf{Rottura della frase}: dato un pezzo di testo vengono trovati i limiti della frase, spesso contrassegnati da punti o altri segni di punteggiatura.
	\item \textbf{Stemming}: alcune parole vengono ridotte alla loro forma radice (ad esempio ``argue, argued, argues, arguing, and argus'' sono mappati alla parola ``argu'').
	\item \textbf{Segmentazione di parole}: un blocco di testo o sentence viene separato in parole. Per una lingua come l'inglese, questo è abbastanza banale, poiché le parole sono solitamente separate da spazi. 
\end{itemize}
Dal punto di vista semantico invece si interviene nel seguente modo:
\begin{itemize}
	\item \textbf{Semantica lessicale}: tenta di comprendere il significato computazionale delle singole parole nel loro contesto.
	\item \textbf{Comprensione del linguaggio naturale}: i blocchi di testo vengono convertiti in rappresentazioni più formali e più facili da manipolare per i computer. 
\end{itemize}
Ricorrendo, dove possibile, alle tecniche sopraelencate, è possibile costruire un dizionario delle \numprint{60000} parole più frequenti nel corpus di training, basato sulla frequenza assoluta di una parola, avendo cura di eliminare tutti gli aggettivi presenti nel dataset OCEAN.

\begin{figure}[H]
	\centering
	{\includegraphics[width=.8\textwidth]{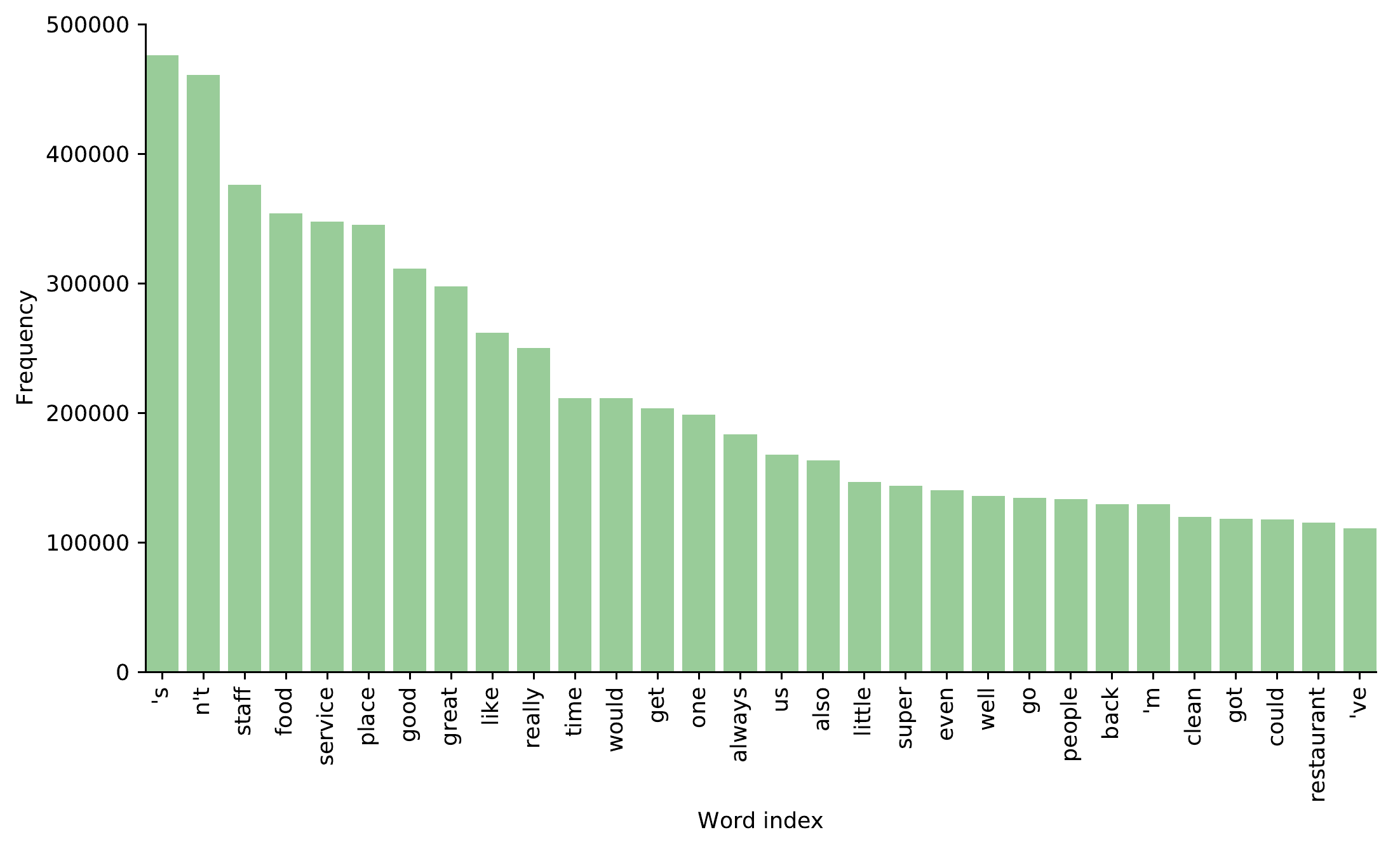}}
	\caption{Istogramma rappresentativo delle 30 parole più frequenti nel dizionario}
	\label{fig:Istrogramma del dizionario}
\end{figure}
In questo modo non verrà influenzata la rete mostrando l'associazione tra gli aggettivi e le label associate, ed il modello sarà ``costretto'' ad imparare il legame esistente tra il contesto di una frase e il relativo valore del tratto di personalità. 

Inoltre sarà necessario anche rimuovere tutte le \emph{stopwords} relative alla lingua inglese, ovvero parole considerate poco significative perché usate troppo frequentemente all'interno delle frasi --- per esempio gli articoli e le congiunzioni --- filtrando i termini comuni e senza uno specifico significato semantico dalle parole che trasportano vere informazioni.

\begin{figure}[H]
	\centering
	{\includegraphics[width=.5\textwidth]{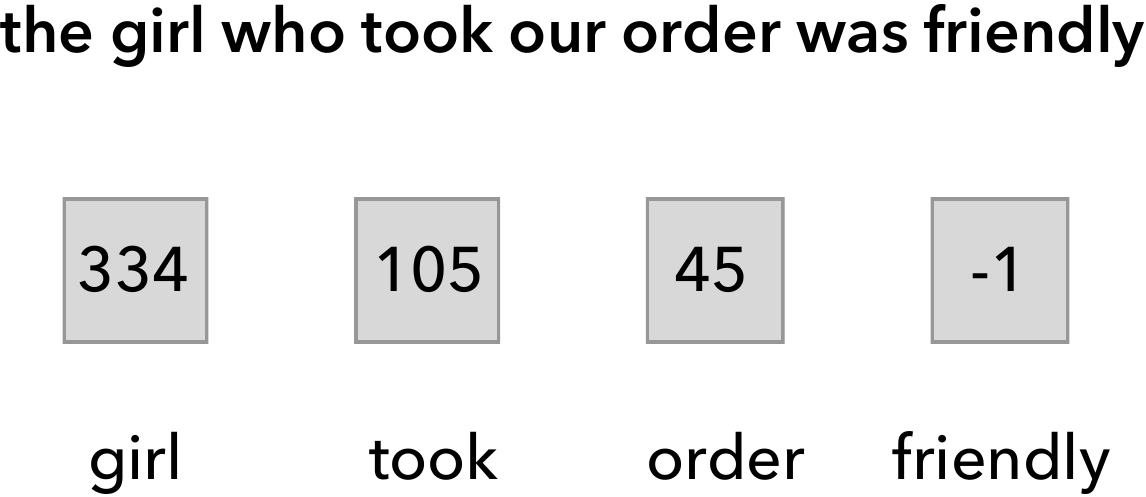}}
	\caption{Visualizzazione del preprocessing}
	\label{fig:preprocessing}
\end{figure}

In seguito ogni parola del dizionario verrà codificata con un valore intero univoco, mentre quelle non presenti, tra cui gli aggettivi mappati nel vocabolario OCEAN, verranno indicizzati al valore ``-1'' e gli verrà associato il token ``UNK'' come mostrato nella Figura~\ref{fig:preprocessing}

\chapter{Esperimenti e risultati}
\label{chap:esperimenti}

La maggior parte degli attuali studi di previsione della personalità si sono concentrati sull'applicazione di tecniche generali di apprendimento automatico per predire i tratti di personalità Big Five.
In particolare verranno utilizzate diverse strutture di rete, combinando differenti domini.

\section{Spazi di rappresentazione}
\label{sec:approcci}
Per affrontare questo problema di \emph{Text Mining}, gli esperimenti si concentrano su due principali metodi per l'estrazione delle caratteristiche del testo:
\begin{itemize}
	\item Un approccio supervisionato, in cui viene utilizzato come strumento di generazione di feature un vettore \emph{bag-of-words} o brevemente BoW, una rappresentazione di testo che descrive la presenza delle parole all'interno di un documento, in questo caso nel dizionario delle occorrenze \cite{wallach2006topic}.
	\item Un approccio non supervisionato, in cui viene costruito un embedding, tramite l'algoritmo \texttt{word2vec} di Tomas Mikolov \cite{mikolov2013distributed}. Insegnando alla rete il significato delle parole e la relazione tra di esse, è possibile rappresentare, sotto forma di vettori, le mappature tra le parole e i contesti.
\end{itemize}
$\\$
In seguito viene posta l'attenzione su tre diverse architetture neurali:
\begin{itemize}
	\item Reti fully-connected;
	\item Reti neurali convoluzionali CNN;
	\item Classificatori multi-label binari.
\end{itemize}

\section{Esperimento 1}
\label{sec:es1}
\subsection{Input Features}
\label{subsec:features1}

Nel primo approccio proposto, per rappresentare i dati testuali viene utilizzato il modello \emph{bag-of-words}, un tipo di descrizione semplificata, spesso utilizzata nell'elaborazione del linguaggio naturale e nel campo dell'\emph{Information Retrivial} (IR). 

Sfruttando il dizionario delle occorrenze precedentemente costruito, il testo viene modellato  come fosse una ``borsa di parole'', in cui grammatica e ordine delle parole vengono trascurate.
Ogni frase viene ridotta ad un vettore in cui ogni elemento identifica una parola del dizionario. Nella posizione corrispondente ad un determinato termine vi sarà il valore 1 se la parola è contenuta nella frase, 0 altrimenti.

Il modello riguarda solo le parole conosciute, di conseguenza i vocaboli che compaiono nel testo ma sono assenti nel dizionario vengono trascurati.

\begin{figure}[H]
	\centering
	{\includegraphics[width=.6\textwidth]{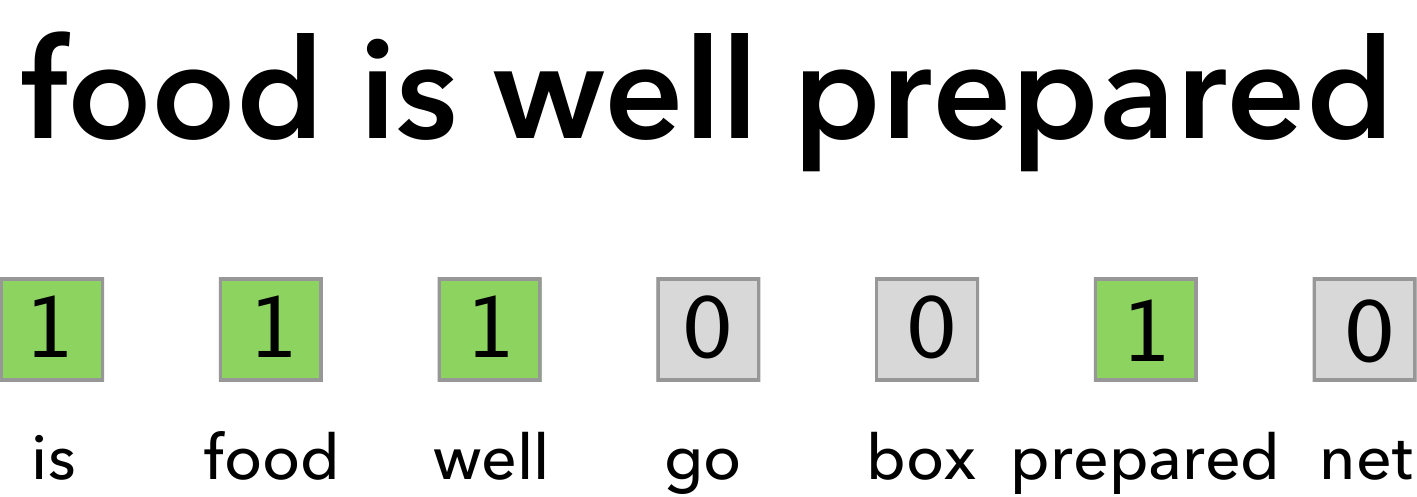}}
	\caption{Visualizzazione del modello \emph{bag-of-words}}
	\label{fig:bow}
\end{figure}

\subsection{Architettura della rete}
\label{subsec:modelli1}

Una volta estratte, le features posso essere passate in input alla rete neurale, che le elabora calcolando le risposte dei neuroni dal livello di input verso il livello di output.

In questa prima strategia viene utilizzata una rete \emph{feed-forward} con struttura densa o \emph{fully-connected}. Per i dettagli di queste architetture si rimanda alla Sezione~\ref{subsec:fc}.

Ad ogni livello viene applicata la funzione di attivazione non lineare \emph{ReLU}, descritta nella Sezione~\ref{subsec:fattivazione}.

Inoltre, per accelerare l'apprendimento ed aumentare la stabilità della rete, viene effettuata dopo ogni layer una \emph{Batch Normalization}, definita nella Sezione~\ref{subsec:normalization}. 

Vengono presentate nella tabella \ref{tab:arcbow+fc} tre architetture implementate, con differenti strati e numero di neuroni che caratterizza ciascuno.

\begin{figure}[H]
	\centering
		\begin{tabular}{lcccc}
			\toprule
			\textbf{Layer} \quad & \textbf{Modello 1} & \textbf{Modello 2} & \textbf{Modello 3} \\
			\midrule
			Input 				 & \numprint{60000}	  & \numprint{60000}   &\numprint{60000}\\
			fc1  				 & \numprint{300}		  & 300 		   & {100} 		\\
			fc2  				 &  {200}		  & 200 		   & 50  		\\
			fc3					 & -				  & {100} 		   & 20    	\\
			Output 				 &  {5}			  & {5} 		   & {5}		\\
			\bottomrule
		\end{tabular}
	\captionof{table}{Confronto delle architetture di tre differenti modelli}
	\label{tab:arcbow+fc}
\end{figure}

Tutte le simulazioni sono state addestrate per {10} epoche: la fase di addestramento viene in genere portata avanti fino a quando le performance sul test non producono alcun miglioramento.

L’ottimizzatore scelto è \emph{Adagrad}, introdotto nella Sezione~\ref{subsubsec:adagrad}, con learning rate \numprint{0,001}.

Come funzione di loss è stato scelto l'errore quadratico medio, in inglese \emph{Mean Squared Error (MSE)}, presentato nella Sezione~\ref{subsubsec:MSE}. Mentre la metrica di valutazione utilizzata per misurare le prestazioni predittive del modello è la \emph{Root Mean Squared Error} (RMSE), introdotta nella Sezione~\ref{subsubsec:regressione}.

In ogni epoca si alternano una fase di training ed una fase di test in modo tale da monitorare costantemente i miglioramenti o i peggioramenti del modello sul test set. 

\subsection{Performance}
\label{subsec:performance1}
Prendendo in considerazione le tre diverse architetture implementate, vengono presentati i risultati ottenuti in termini di loss.
\begin{table}[H]
	\centering
	\begin{tabular}{l@{\hspace{.5cm}}ccc}
		\toprule
		 & \textbf{Train loss} & \textbf{Test loss} & \textbf{Tempo di training}  \\
		\midrule
		\textbf{Modello 1} & \numprint{0.061} & \numprint{0.062} &\numprint{235} min \\
		\textbf{Modello 2} & \numprint{0.090} & \numprint{0.061} &\numprint{250} min \\
		\textbf{Modello 3} & \numprint{0.068} & \numprint{0.062} &\numprint{265} min \\
		\bottomrule 
	\end{tabular}
	\captionof{table}{Confronto dei risultati in termini di {loss} ottenuti dalle tre diverse reti}
	\label{tab:lossbow+fc}
\end{table}

Per valutare l'efficacia di questi modelli, è fondamentale eseguire un analisi dettagliata, in particolare ponendo l'attenzione sui valori di RMSE per ogni tratto di personalità.
Calcolando il valore medio assunto da ogni caratteristica durante la fase di addestramento, è possibile stabilire qual è il valore di Root Mean Squared Error di un modello concettuale, chiamato ``Modello 0'', che per ogni tratto predice sempre il suo valore medio.
 
\begin{figure}[H]
	\centering
	\begin{tabular}{clccccc}
		\toprule	
		& 		  & \multicolumn{5}{c}{\textbf{Root Mean Squared Error}} 									    \\
		\multicolumn{2}{c}{\multirow{-2}{*}{Modelli}}
		& O 				& C 			   & E 				  & A 				 & N 			    \\ 
		\midrule
		\multirow{2}*{\textbf{Modello 1}} & Modello   & \numprint{0,148} & \numprint{0,227} & \numprint{0,224} & \numprint{0,251} & \numprint{0,351} \\
		& Modello 0 & \numprint{0,145} & \numprint{0,224} & \numprint{0,213} & \numprint{0,218} & \numprint{0,318} \\
		\midrule
		\multirow{2}*{\textbf{Modello 2}} & Modello   & \numprint{0,147} & \numprint{0,226} & \numprint{0,225} & \numprint{0,251} & \numprint{0,341} \\
		& Modello 0 & \numprint{0,141} & \numprint{0,227} & \numprint{0,213} & \numprint{0,208} & \numprint{0,305} \\
		\midrule
		\multirow{2}*{\textbf{Modello 3}} & Modello   & \numprint{0,147} & \numprint{0,226} & \numprint{0,225} & \numprint{0,262} & \numprint{0,348} \\
		& Modello 0 & \numprint{0,233} & \numprint{0,307} & \numprint{0,262} & \numprint{0,373} & \numprint{0,546}  \\
		\bottomrule	
	\end{tabular}
	\captionof{table}{Confronto dei risultati in termini di {Root Mean Squared Error} delle architetture contro il modello basato sulla media di training}
	\label{tab:confmm0bow+fc}
\end{figure}

\begin{figure}[htb]
	\centering
	{\includegraphics[width=.75\textwidth]{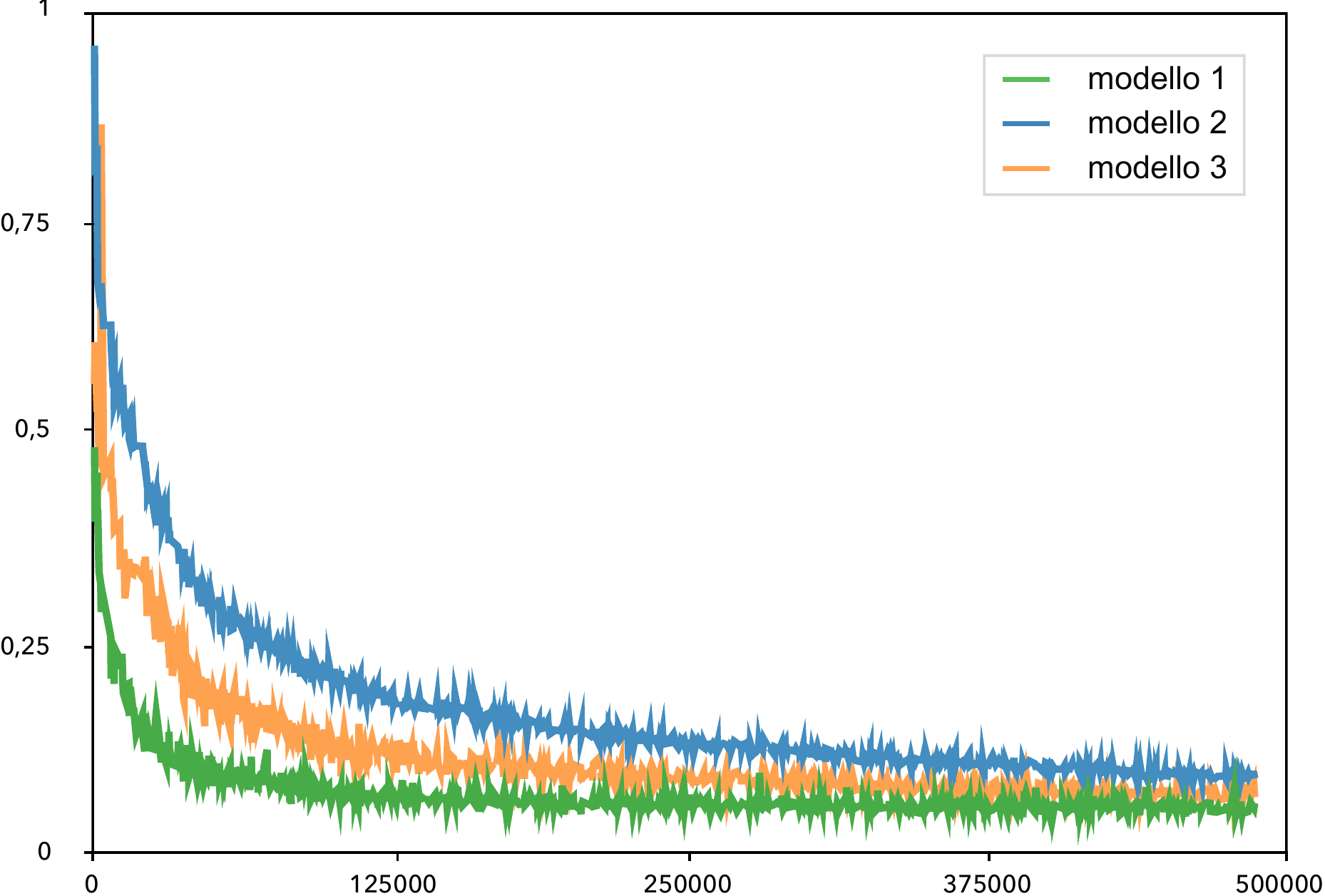}} 
	\caption{Visualizzazione delle training loss dei tre modelli}
	\label{fig:loss}
\end{figure}

Mettendo a confronto le due metriche, si nota che i modelli realmente implementati imparano nella maggioranza dei casi a predire un valore con lo stesso errore commesso dai modelli banali che calcolano la media. Risulta allora evidente che i risultati ottenuti non siano ottimali.

Nonostante ciò, il terzo modello sembrerebbe essere leggermente migliore degli altri due, in particolare confrontandolo con il corrispondente modello nullo si nota la presenza di un piccolo margine di miglioramento.

La scarsa efficienza di questi modelli potrebbe dipendere dall'efficacia con cui vengono codificate le feature in input alla rete. 
Infatti le limitazioni dell'approccio bag-of-words derivano in parte dalla progettazione del vocabolario e della sua dimensione, che può causare una scarsa ``descrizione'' del documento. 
Scartare l'ordine delle parole e ignorare il contesto non consente di determinare la differenza tra le stesse parole disposte diversamente, i sinonimi ecc.
Inoltre, un tipo di rappresentazione sparsa risulta più difficile da modellare quando si cercano modelli in grado di sfruttare poche informazioni in uno spazio rappresentativo ampio, sia per ragioni computazionali (spazio e complessità temporale) sia per ragioni di informazione.

\section{Esperimento 2}
\label{sec:es2}

Durante l'applicazione di tecniche di apprendimento automatico, tutte le ``informazioni'' vengono rappresentate per mezzo di identificativi unici e discreti. 

Nel caso dell'approccio {BoW}, la codifica utilizzata non fornisce alcuna informazione utile al sistema riguardo le relazioni che possono sussistere tra i singoli elementi. Ciò significa che quando sta elaborando i dati, il modello può sfruttare molto poco di ciò che ha appreso su un determinato termine. 

Inoltre la ``raffigurazione'' utilizzata nel precedente esperimento, ha portato alla creazione di dati sparsi. Di conseguenza, per ottenere un modello di successo, un'alternativa valida sarebbe quella di sfruttare {modelli spaziali vettoriali}, in inglese \emph{Vector Space Model} (VSM), per rappresentare le parole in uno spazio continuo \cite{mikolov2013linguistic,erk2008structured}.
Questi metodi dipendono dall'ipotesi distributiva, la quale afferma che le parole che appaiono negli stessi contesti condividono lo stesso significato semantico \cite{baroni2014don}. 

\subsection{Input Features}
\label{subsec:features2}

Nel secondo approccio, viene applicato l'algoritmo non supervisionato \texttt{word2vec} di Tomas Mikolov  \cite{mikolov2013efficient}. 
\texttt{Word2vec} è un modello predittivo particolarmente efficiente dal punto di vista computazionale per l'apprendimento degli embedding di parole a partire dal testo non elaborato.
Esso è basato su una rete neurale artificiale a due strati, addestrati a ricostruire i contesti linguistici delle parole. 

A partire dal corpus di testo, la rete prende in input un set formato dall'accoppiamento di ogni parola target e i contesti in cui appare e restituisce un insieme di vettori che rappresentano la distribuzione semantica delle parole nel testo. 

Viene considerato come ``contesto'' l'insieme delle ``parole a sinistra'' e delle ``parole alla destra'' dell'obiettivo, ovvero la finestra di dimensione 1 attorno all'elemento target. Ogni coppia di destinazione del contesto viene trattata come se fosse una nuova osservazione, incrementando le informazioni distribuzionali. Viene così prodotto uno spazio vettoriale di diverse centinaia di dimensioni, in cui ogni parola univoca viene assegnata a un vettore corrispondente nello spazio.

Per l'implementazione viene utilizzato il modello \emph{skip-gram}, una versione di \texttt{word2vec} che vuole predire le parole del contesto di origine (label) a partire dalle parole target (features).

\begin{figure}[H]
	\centering
	{\includegraphics[width=.7\textwidth]{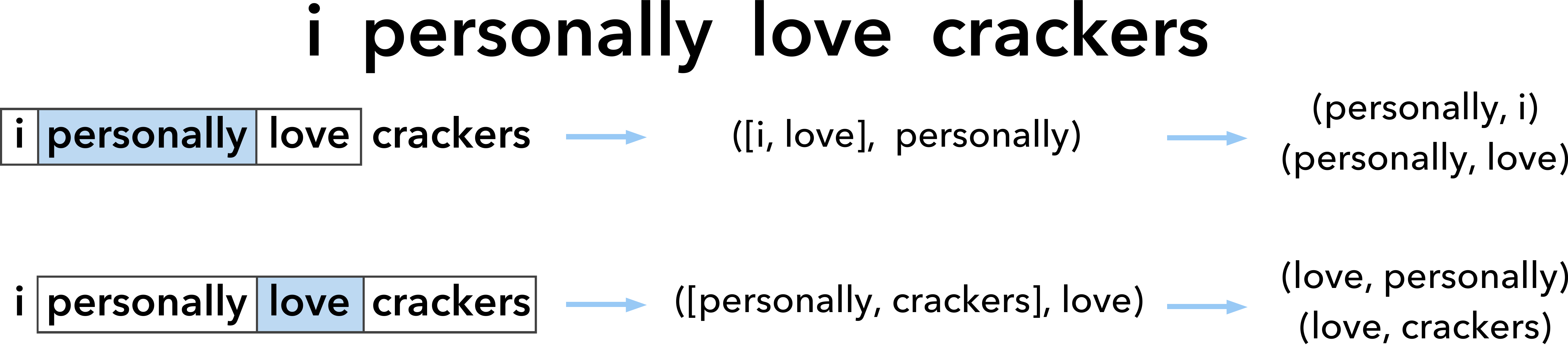}} 
	\caption{Visualizzazione del modello skip-gram}
	\label{fig:mikolov}
\end{figure}

\begin{figure}[H]
	\centering
	\subfloat[][\emph{Visualizzazione 2D}\label{subfig:vis2d}]
	{\includegraphics[width=.4\textwidth]{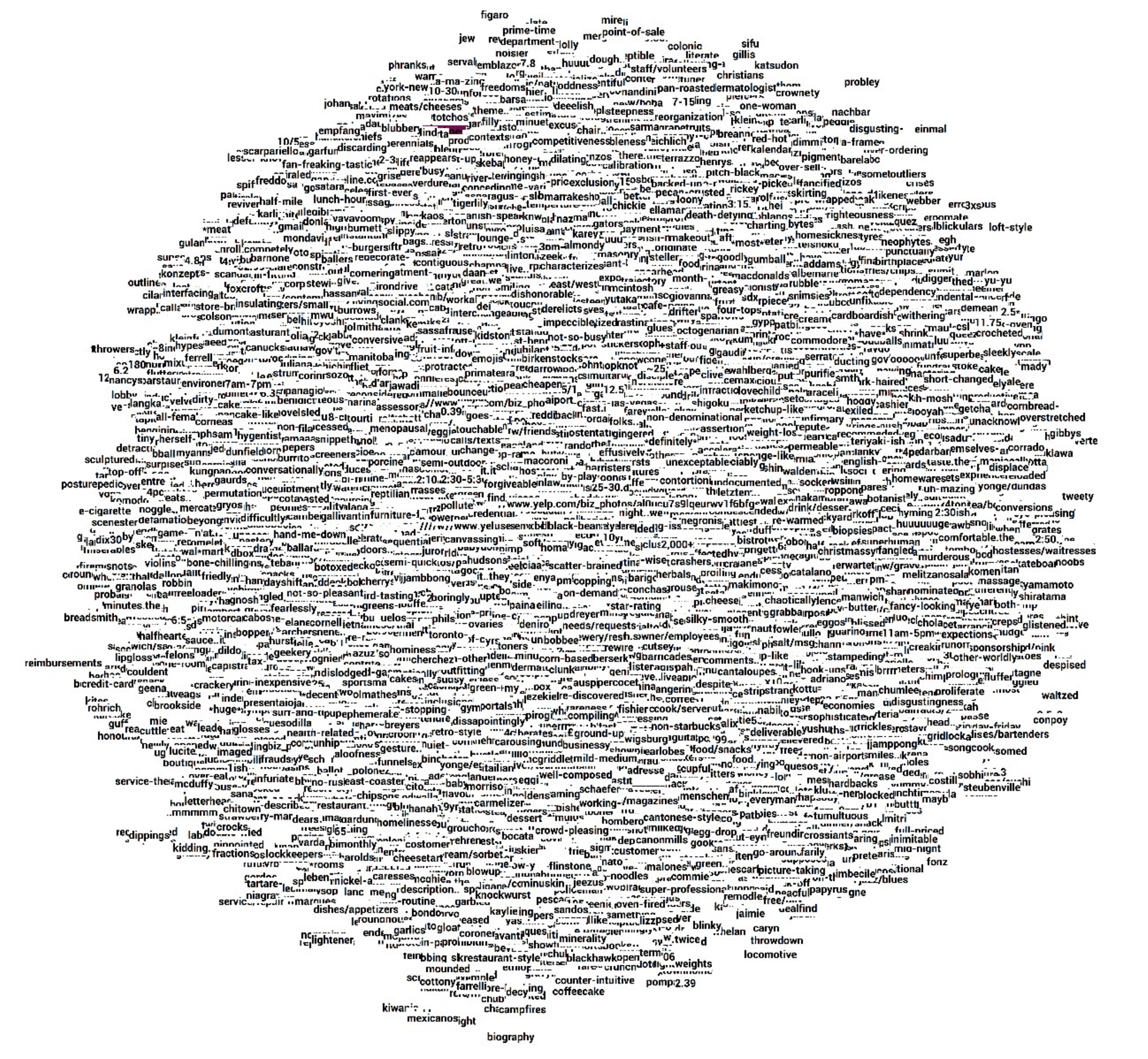}}
	\hspace{10mm}
	\subfloat[][\emph{Dettaglio di una visualizzazione 3D}\label{subfig:vis3d}]
	{\includegraphics[width=.45\textwidth]{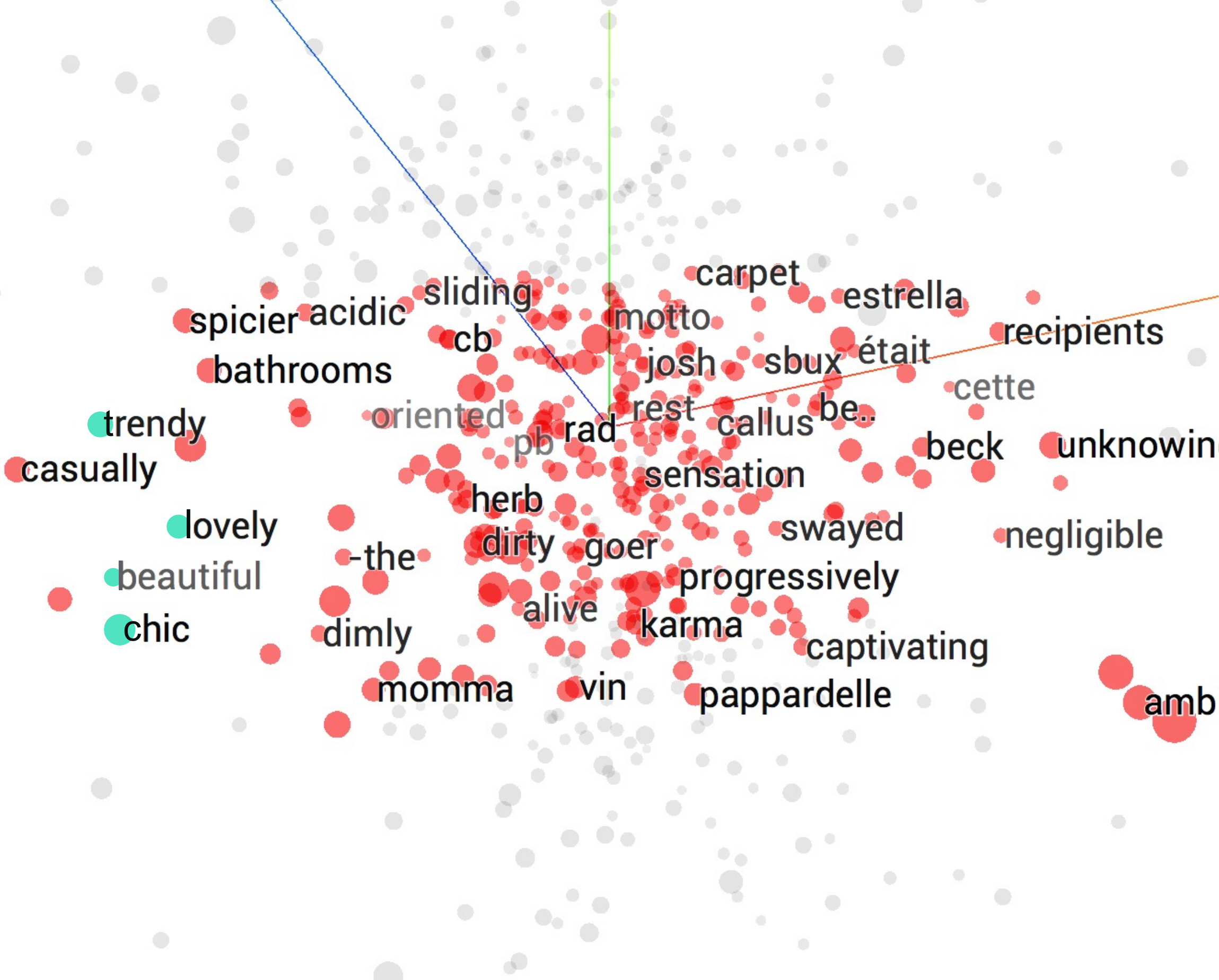}}
	
	\caption{Proiezione dell'embedding di Mikolov nello spazio 2-3 dimensionale, tramite la \emph{tecnica di riduzione della dimensionalità} (t-SNE) \cite{maaten2008visualizing}}
	\label{fig:embedding1}
\end{figure}

In questo tipo di apprendimento, i vettori si posizionano nello spazio in modo tale che le parole che condividono contesti comuni nel corpo siano situate in stretta prossimità l'una dell'altra. Un esempio interessante viene illustrato nella Figura~\ref{fig:embedding1}\subref{subfig:vis3d} --- si ponga particolare attenzione alle parole ``beautiful'', ``lovely'', ``chic'', ``trendy'' ecc.. --- in cui i vocaboli semanticamente simili si trovano vicini dello spazio.

La funzione obiettivo utilizzata dalla rete per la costruzione dell'embedding viene definita sull'intero set di dati, ed ottimizzata con la \emph{Stochastic Gradient Descent} (SGD), definita nella Sezione~\ref{subsubsec:SGD}.

Nella seguente tabella vengono presentati i due embedding realizzati e i relativi parametri.

\begin{figure}[htb]
	\centering
	\begin{tabular}{ccc}
		\toprule	
		 		  				& \multicolumn{2}{c}{\textbf{Parameters}}	\\
		{\multirow{-2}{*}{Embedding}}
								& Embedding Size 	& Num Sampled 	 		\\ 
		\midrule
		\textbf{Embedding 1}    & \numprint{40} 	& \numprint{20}  		\\
		\midrule
		\textbf{Embedding 2}    & \numprint{250} 	& \numprint{50}  		\\
		\bottomrule	
	\end{tabular}
	\captionof{table}{Confronto dei parametri della rete impostati per la realizzazione dell'embedding}
	\label{tab:confemb}
\end{figure}

\subsection{Architettura della rete}
\label{subsec:modelli2}

La rappresentazione spazio-vettoriale viene utilizzata come feature della modello che si andrà a costruire.
In questo secondo approccio verrà utilizzata un \emph{rete convoluzionale}, in cui ogni neurone è collegato solo a pochi neuroni vicini nel livello precedente, e lo stesso insieme di pesi viene utilizzato per ogni neurone.\\

In questo tipo di rete, il modello di connessione locale e lo schema di peso condiviso possono essere interpretati come un filtro (o un insieme di filtri) che accettano un sottoinsieme dei dati di input alla volta, ma vengono applicati all'intero input.

Uno strato convoluzionale è molto più specializzato ed efficiente di uno completamente connesso.

\begin{figure}[H]
	\centering
	{\includegraphics[width=.85\textwidth]{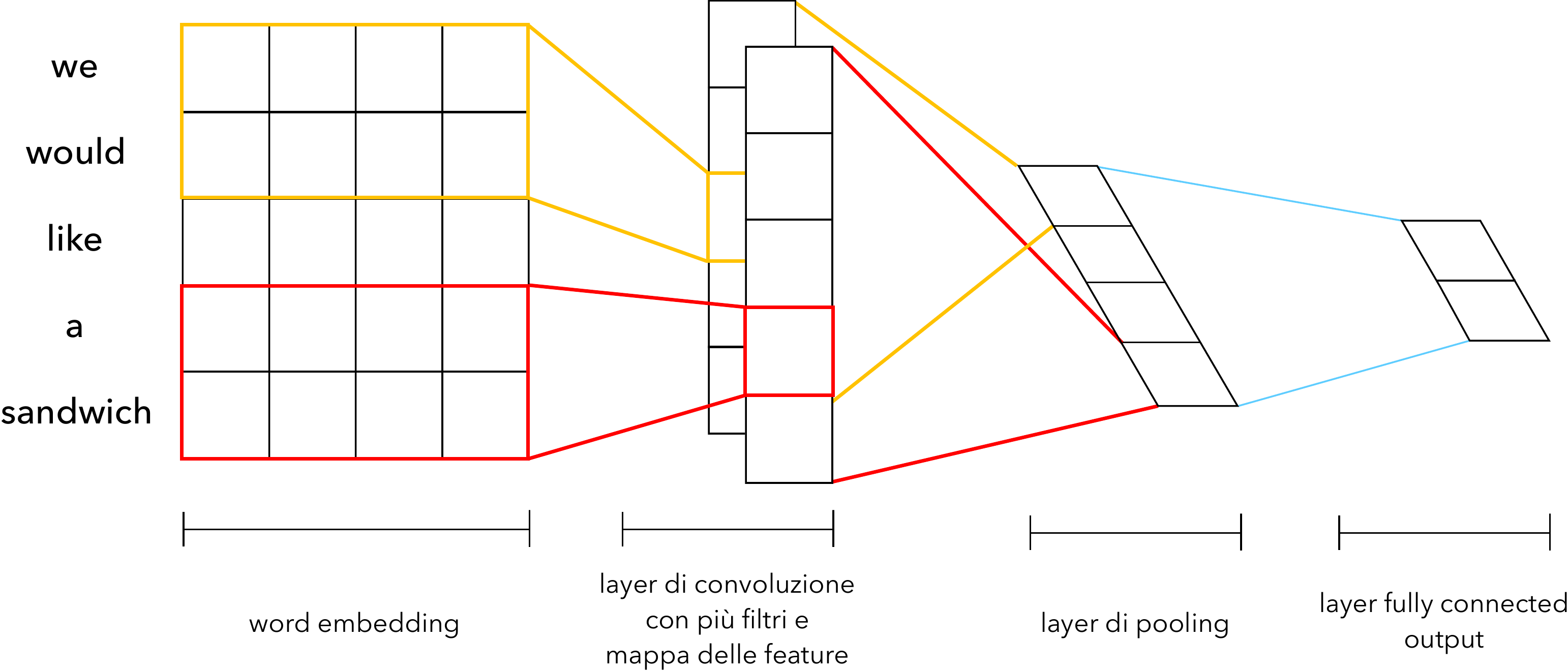}} 
	\caption{Esempio di architettura convoluzionale per la manipolazione del linguaggio naturale}
	\label{fig:cnn}
\end{figure}

Le operazioni eseguite da questi strati vengono trasformate in ``moltiplicazioni'' non lineari tramite l'applicazione della funzione di attivazione \emph{ReLU}, introdotta nella Sezione~\ref{subsec:fattivazione}.

Ad ogni livello convoluzionale viene applicata una \emph{Batch Normalization}, definita nella Sezione~\ref{subsec:normalization}, utilizzata sull'input per il ridimensionamento delle funzionalità e la normalizzazione batch nei livelli nascosti.

Dopo il primo strato convoluzionale viene inserito un \emph{layer di pooling}, introdotto nella Sezione~\ref{subsec:maxpool}, necessario per ridurre in modo efficace i campioni dell'output del livello precedente, riducendo il numero di operazioni richieste per tutti i livelli successivi, ma passando comunque le informazioni valide.

Come ultimo strato della rete viene scelto un \emph{layer fully-connected}, definito nella Sezione~\ref{subsec:fc}, corrispondente ad un'operazione lineare sul vettore di input del livello che esegue una serie di trasformazioni sulla rappresentazione profonda al fine di emettere i punteggi di ogni classe.

\begin{figure}[H]
	\centering
	\begin{tabular}{lccc}
		\toprule
		\textbf{Layer}& \textbf{Modello 4} & \textbf{Modello 5} & \textbf{Modello 6} 		\\ 
		\midrule
		conv1 	& {10}$\times${5}, 10	  & {5}$\times${5}, 150, same pad    &{3}$\times${3}, 100, same pad 		   \\
		
		mpool1 	& &{{4}$\times${4}, stride 2, same pad}	&   \\
		conv2  	& {18}$\times${18}, 10	  &  {5}$\times${20}, 100, same pad	  &		{3}$\times${20}, 75, same pad    \\
		conv3  	& ---	  & {1}$\times${20}, 50, 	   &	{1}$\times${20}, 50 	   \\
		fcout		& &{{1}$\times${1}, 5}&		   \\
		
		\bottomrule	
	\end{tabular}
	\captionof{table}{Architetture implementate a partire dall'embedding 1}
	\label{tab:netemb1}
\end{figure}

\begin{figure}[H]
	\centering
	\begin{tabular}{lcc}
		\toprule
		\textbf{Layer}& \textbf{Modello 7} 								  & \textbf{Modello 8} 			   \\ 
		\midrule
		conv1 	& {3}$\times${3}, 100, stride 2, same pad     & ---	   \\
		mpool1 	& {4}$\times${4}, stride 2, same pad		  & ---	   \\
		conv2  	& {3}$\times${63}, 75, stride 2, same pad	  & ---    \\
		conv3  	& {1}$\times${32}, 50, stride 2	  				  & ---	   \\
		fc1  	& ---													  & {1}$\times${1}, 100	   \\
		fc2  	& {1}$\times${32}, 50, stride 2	  				  & {1}$\times${1},  50    \\
		fc3  	& {1}$\times${32}, 50, stride 2	  				  & {1}$\times${1},  20	   \\
		fcout	& {1}$\times${1}, 5   			  				  & {1}$\times${1},   5	   \\
		\bottomrule	
	\end{tabular}
	\captionof{table}{Architetture implementate a partire dall'embedding 2}
	\label{tab:netemb2}
\end{figure}

Vengono presentate nelle tabelle \ref{tab:netemb1} e \ref{tab:netemb2} le diverse configurazioni dei layer convoluzionali delle architetture implementate per i due embedding.\\

L'input della rete è costituito da un numero di caratteristiche pari a $n \times p$ dove $n$ è la dimensione dell'embedding e $p$ il numero delle parole di ogni sentence.

Come algoritmo di ottimizzazione viene scelto sempre \emph{Adagrad}, illustrato nella Sezione~\ref{subsubsec:adagrad}, mentre il valore di \emph{learning rate} settato per ogni modello viene mostrato nella tabella \ref{tab:learningratemikolov}.

Viene mantenuta anche in questa simulazione la funzione di costo MSE, approfondita nella Sezione~\ref{subsubsec:MSE}.

Nel caso del secondo embedding si è voluto provare anche un modello che sfruttasse solo livelli di rete \emph{fully-connected}, senza applicare alcuna convoluzione.

\begin{table}[t]
	\centering
	\begin{tabular}{llc}
		\toprule
		\multicolumn{2}{c}{{Modelli}} & \textbf{Learning Rate}  \\
		\midrule
		\multirow{3}*{{Embedding 1}} 
		&\textbf{Modello 4} & \numprint{0.0010} \\
		&\textbf{Modello 5} & \numprint{0.0001} \\
		&\textbf{Modello 6} & \numprint{0.0050} \\
		\midrule
		\multirow{2}*{{Embedding 2}} 
		&\textbf{Modello 7} & \numprint{0.0050} \\
		&\textbf{Modello 8} & \numprint{0.0001} \\	
		\bottomrule 
	\end{tabular}
	\captionof{table}{Learning rate delle simulazioni effettuate nell'esperimento 2}
	\label{tab:learningratemikolov}
\end{table}

\subsection{Performance}
\label{subsec:performance2}

Prendendo in considerazione i due diversi embedding, vengono messe a confronto a confronto le diverse architetture implementate, mostrando i risultati ottenuti in termini di loss.
\begin{table}[H]
	\centering
	\begin{tabular}{ll@{\hspace{.5cm}}ccc}
		\toprule
		\multicolumn{2}{c}{{Modelli}} & \textbf{Train loss} & \textbf{Test loss} & \textbf{Tempo di training}  \\
		\midrule
		\multirow{3}*{{Embedding 1}} 
		&\textbf{Modello 4} & \numprint{0.061} & \numprint{0.058} &\numprint{200} min \\
		&\textbf{Modello 5} & \numprint{0.052} & \numprint{0.060} &\numprint{310} min \\
		&\textbf{Modello 6} & \numprint{0.042} & \numprint{0.060} &\numprint{540} min \\
		\midrule
		\multirow{2}*{{Embedding 2}} 
		&\textbf{Modello 7} & \numprint{0.038} & \numprint{0.057} &\numprint{225} min \\
		&\textbf{Modello 8} & \numprint{0.058} & \numprint{0.117} &\numprint{250} min \\	
		\bottomrule 
	\end{tabular}
	\captionof{table}{Confronto dei risultati in termini di \emph{loss} ottenuti nelle diverse reti con i due diversi embedding}
	\label{tab:lossmikolov}
\end{table}

Valutando le prestazioni su train e test set viene rilevato il ``Modello 7'' come il migliore tra quelli proposti, presentando i valori di loss più bassi. 
Oltre a ciò, anche rispetto al precedente approccio il ``Modello 7'' mostra una performance migliorata circa 
del $3\%$.

\begin{figure}[H]
	\centering
	{\includegraphics[width=.65\textwidth]{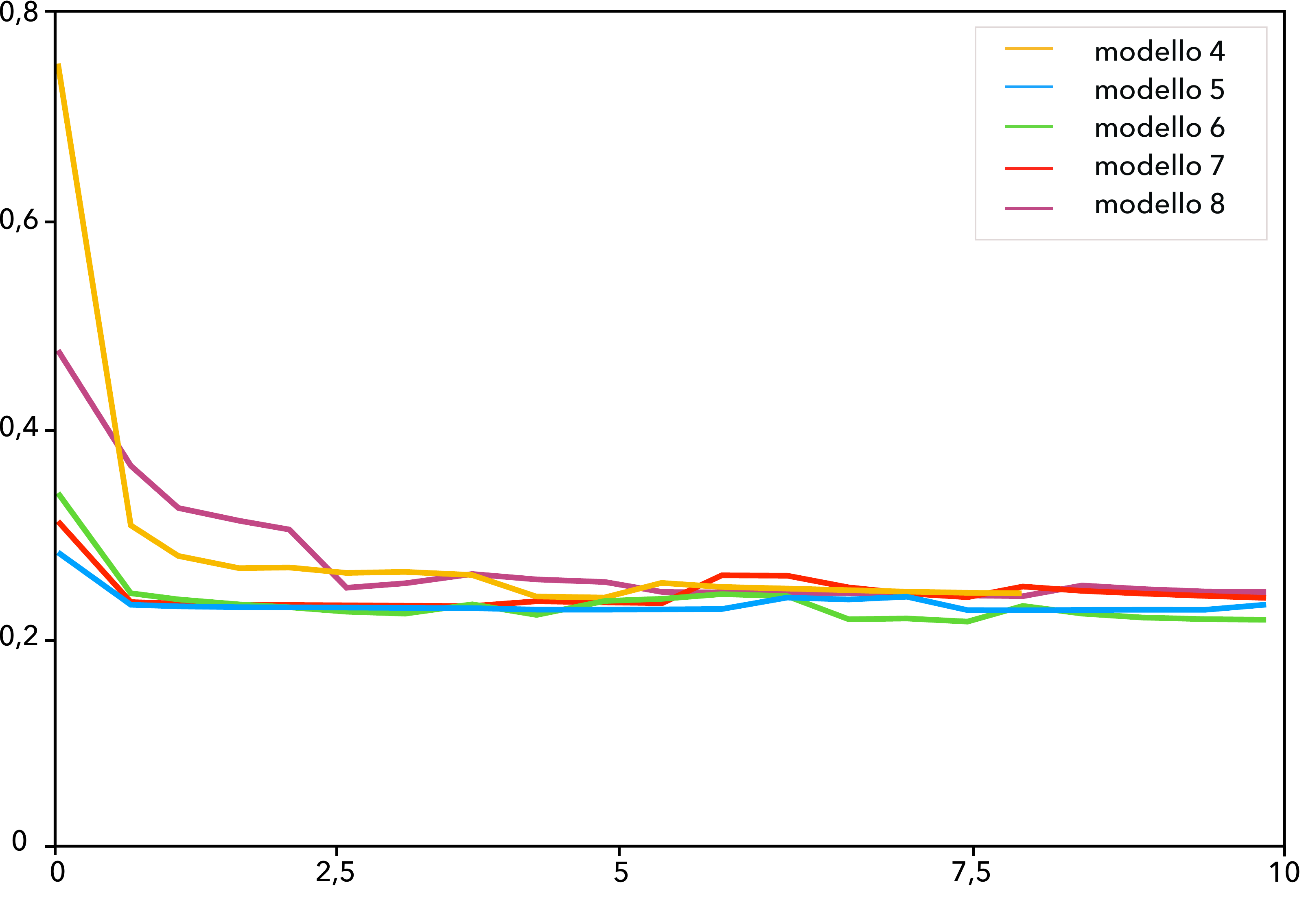}} 
	\caption{Visualizzazione delle training RMSE dei modelli}
	\label{fig:rmse}
\end{figure}

Vengono analizzati anche i valori di RMSE, relativi ad ogni tratto di personalità, riassunti nella tabella \ref{tab:rmsemikolov}, e viene inoltre visualizzata la curva RMSE nella Figura~\ref{fig:rmse}.

\begin{figure}[t]
	\centering
	\begin{tabular}{clccccc}
		\toprule	
		& 		 			& \multicolumn{5}{c}{\textbf{Root Mean Squared Error}} 									       \\
		\multicolumn{2}{c}{\multirow{-2}{*}{Modelli}}
		& O 				& C 			   & E 				  & A 				 & N 			   \\ 
		\midrule
		\multirow{3}*{{Embedding 1}} 
		& \textbf{Modello 4} & \numprint{0,148} & \numprint{0,226} & \numprint{0,232} & \numprint{0,252} & \numprint{0,313} \\
		& \textbf{Modello 5} & \numprint{0,146} & \numprint{0,227} & \numprint{0,230} & \numprint{0,251} & \numprint{0,336} \\
		& \textbf{Modello 6} & \numprint{0,146} & \numprint{0,225} & \numprint{0,224} & \numprint{0,251} & \numprint{0,337} \\
		\midrule
		\multirow{2}*{{Embedding 2}} 
		& \textbf{Modello 7} & \numprint{0,147} & \numprint{0,223} & \numprint{0,222} & \numprint{0,251} & \numprint{0,320} \\
		& \textbf{Modello 8} & \numprint{0,399} & \numprint{0,275} & \numprint{0,257} & \numprint{0,271} & \numprint{0,457} \\
		\bottomrule	
	\end{tabular}
	\captionof{table}{Confronto dei risultati in termini di {Root Mean Squared Error} delle architetture sul test set}
	\label{tab:rmsemikolov}
\end{figure}

\section{Esperimento 3}
\label{sec:es3}

Nel terzo approccio si decide di valutare una rappresentazione dell'input alternativa.

\subsection{Input Features}
\label{subsec:features3}

Ricorrendo nuovamente al modello \emph{skip-gram}, l'input della rete questa volta comprende il set formato dall'accoppiamento tra gli aggettivi contenuti nel dizionario OCEAN e i loro contesti.

La finestratura che si andrà a definire in torno al target sarà di dimensione 2 e considererà le due parole a sinistra e le due parole a destra dell'aggettivo. La window non sarà più quindi incentrata su ogni elemento della sentence.
Dunque nell'embedding che si andrà a costruire non verranno appresi i contesti di tutte le parole, ma solamente quelli di nostro interesse, ovvero relativi agli aggettivi.

In questo esperimento viene realizzato un solo embedding, la cui dimensione è {250} e con numero di etichette negative da campionare pari a 50 \cite{liu2016classification}.

\subsection{Architettura della rete}
\label{subsec:modelli3}

Viene utilizzato l'embedding estratto per costruire due modelli di reti neurali, le cui architetture vengono presentate nella tabella \ref{tab:netemb3}.
Come nel precedente esperimento si ricorrerà alle \emph{reti convoluzionali}; si rimanda alla Sezione~\ref{subsec:modelli2} per i dettagli. 

\begin{figure}[H]
	\centering
	\begin{tabular}{lcc}
		\toprule
		\textbf{Layer}& \textbf{Modello 9} & \textbf{Modello 10}	\\ 
		\midrule
		conv1   & {7}$\times${5}, 100, stride 2, same pad    		&{3}$\times${3}, 100, stride 2, same pad 		   \\
		mpool1 	&\multicolumn{2}{c}{{4}$\times${4}, stride 2, same pad}	 \\
		conv2  	&  {5}$\times${63}, 75, stride 2, same pad	    &		{3}$\times${63}, 75, stride 2, same pad    \\
		conv3  	& {3}$\times${32}, 50, stride 2, same pad 	   	&	{1}$\times${32}, 50, stride 2 	   \\
		conv4  	& {1}$\times${16}, 25, stride 2 	   	&	--- 	   \\
		fcout	&\multicolumn{2}{c}{{1}$\times${1}, 5}						\\
		
		\bottomrule	
	\end{tabular}
	\captionof{table}{Architetture dei due modelli implementati a partire dall'embedding 3}
	\label{tab:netemb3}
\end{figure}

L'algoritmo di apprendimento utilizzato è \emph{Adagrad}, definito nella Sezione~\ref{subsubsec:adagrad}, con \emph{learning rate} \numprint{0,0005} per il ``Modello 9'' e \numprint{0,005} per il ``Modello 10''. La funzione di \emph{loss} scelta è la MSE, introdotta nella Sezione~\ref{subsubsec:MSE}. 

\subsection{Performance}
\label{subsec:performance3}

Vengono presentati nella tabella \ref{tab:lossmikolov2} i valori ottenuti dalla funzione obiettivo dei due modelli, mentre nella tabella \ref{tab:rmsemikolov2} vengono riassunti i valori di RMSE per ogni tratto di personalità.

\begin{table}[H]
	\centering
	\begin{tabular}{l@{\hspace{.5cm}}ccc}
		\toprule
		& \textbf{Train loss} & \textbf{Test loss} & \textbf{Tempo di training}  \\
		\midrule
		\textbf{Modello 9} & \numprint{0.043} & \numprint{0.060} &\numprint{18} h \\
		\textbf{Modello 10} & \numprint{0.050} & \numprint{0.059} &\numprint{17} h \\	
		\bottomrule 
	\end{tabular}
	\captionof{table}{Confronto dei risultati in termini di {loss} ottenuti nell'implementazione dei due modelli}
	\label{tab:lossmikolov2}
\end{table}

\begin{figure}[H]
	\centering
	\begin{tabular}{clccccc}
		\toprule	
		& 		 			& \multicolumn{5}{c}{\textbf{Root Mean Squared Error}} 									       \\
		\multicolumn{2}{c}{\multirow{-2}{*}{Modelli}}
		& O 				& C 			   & E 				  & A 				 & N 			   \\ 
		\midrule
		& \textbf{Modello 9} & \numprint{0,146} & \numprint{0,223} & \numprint{0,223} & \numprint{0,251} & \numprint{0,331} \\
		& \textbf{Modello 10} & \numprint{0,147} & \numprint{0,225} & \numprint{0,224} & \numprint{0,252} & \numprint{0,339} \\
		\bottomrule	
	\end{tabular}
	\captionof{table}{Confronto dei risultati in termini di {RMSE} delle due architetture sul test set}
	\label{tab:rmsemikolov2}
\end{figure}

Rispetto al precedente approccio non vengono ottenuti dei miglioramenti. 

\section{Esperimento 4}
\label{sec:es4}

L'ultimo approccio provato riutilizza i precedenti metodi di estrazione di feature che hanno ottenuto le migliori prestazioni per realizzare un modello predittivo di classificazione.

Nello specifico il problema viene trasformato in un compito di classificazione binaria multi-label, in cui si cercano di predire cinque diverse etichette per ogni istanza. 

Per ogni dimensione di personalità l'output della rete sarà ``0'', se il valore reale che assume un determinato tratto è inferiore a 0, o ``1'' altrimenti.
\\
Viene stabilito 0 come ``punto di neutralità assoluta'' perché corrisponde alla media dei valori osservati in ognuna delle caratteristiche.

Adottando questo metodo, si assumerà di poter estrarre da questa funzione una misura di polarità, positiva o negativa, che indicherà un tratto più o meno accentuato.

Il vantaggio effettivo di questo approccio consiste nella possibilità di valutare le performance in modo standard, visualizzando le matrici di confusione e misurandone l'accuratezza, argomenti trattati nella Sezione~\ref{subsubsec:classificazione}.

\subsection{Input Features}
\label{subsec:features4}

Vengono utilizzati come ingresso della rete due embedding di Mikolov realizzati negli esperimenti precedenti, ovvero l'``Embedding 2'' e l'``Embedding 3''.

\subsection{Architettura della rete}
\label{subsec:modelli4}

Le reti che vengono realizzate sono molto simili a quelle precedentemente implementate; i dettagli delle architetture vengono presentati nella tabella \ref{tab:netemb4}.

Ad ogni livello della rete viene applicata la stessa procedura degli esperimenti precedenti. 

\begin{figure}[H]
	\centering
	\begin{tabular}{lccc}
		\toprule
		\textbf{Layer} & \textbf{Modello 11} & \textbf{Modello 12} & \textbf{Modello 13}  	\\ 
		\midrule
		conv1 	& {3}$\times${3}, 100	  & {5}$\times${3}, 100 & {7}$\times${5}, 100 	   \\
		mpool1 	& {{4}$\times${4}}	&{{4}$\times${4}}	& {{4}$\times${4}}	 \\
		conv2  	& {3}$\times${63}, 75 &  {3}$\times${63}, 75	  &		{5}$\times${63}, 75    \\
		conv3  	&{1}$\times${32}, 50	  & {1}$\times${32}, 50  &	{3}$\times${32}, 50 	  \\
		conv4  	& ---	  & {1}$\times${16}, 25  &	{1}$\times${16}, 25 	  \\
		fcout	&{{1}$\times${1}, 10} &{{1}$\times${1}, 10}&{{1}$\times${1}, 10}		   \\
		\bottomrule	
	\end{tabular}
	\captionof{table}{Architetture implementate a partire dall'embedding 2}
	\label{tab:netemb4}
\end{figure}

In ogni livello convoluzionale della rete si utilizzano uno stride di dimensione 2 e padding same, tranne che per l'ultimo livello convoluzionale con padding valid, per i dettagli si consiglia di visitare la sezione { \ref{subsec:cnn}}.

\begin{figure}[H]
	\centering
	\begin{tabular}{lcc}
		\toprule
		\textbf{Layer} & \textbf{Modello 14} & \textbf{Modello 15}   	\\ 
		\midrule
		conv1 	& {{7}$\times${5}, 100, stride 2, same pad}&{{7}$\times${5}, 100, stride 2, same pad} \\
		mpool1 	& {{4}$\times${4}, stride 2, same pad} &{{4}$\times${4}, stride 2, same pad}  \\
		conv2  	& {{5}$\times${63}, 75, stride 2, same pad}&{{5}$\times${63}, 75, stride 2, same pad}    \\
		conv3  	&  {{3}$\times${32}, 50, stride 2, same pad} &{{3}$\times${32}, 50, stride 2, same pad} 	  \\
		conv4  	& {1}$\times${16}, 25, stride 2  &	{3}$\times${16}, 25, stride 2, same pad 	  \\
		conv5  	& ---  &	{1}$\times${8}, 16, stride 2 	  \\
		fcout	& \multicolumn{2}{c}{{1}$\times${1}, 10}		   \\
		\bottomrule	
	\end{tabular}
	\captionof{table}{Architetture implementate a partire dall'embedding 3}
	\label{tab:rmsebin2}
\end{figure}

In tutte le simulazioni l’ottimizzatore scelto è \emph{Adagrad}, introdotto nella Sezione~\ref{subsubsec:adagrad}, il valore di \emph{learning rate}  di ogni modello viene mostrato nella tabella \ref{tab:learningratemikolov2}.

Come funzione obiettivo viene utilizzata la \emph{Softamax Cross Entropy}, presentata nella Sezione~\ref{subsubsec:sce}. 

\begin{table}[H]
	\centering
	\begin{tabular}{llc}
		\toprule
		\multicolumn{2}{c}{{Modelli}} & \textbf{Learning Rate}  \\
		\midrule
		\multirow{3}*{{Embedding 2}} 
		&\textbf{Modello 11} & \numprint{0.0001} \\
		&\textbf{Modello 12} & \numprint{0.0005} \\
		&\textbf{Modello 13} & \numprint{0.0005} \\
		\midrule
		\multirow{2}*{{Embedding 3}} 
		&\textbf{Modello 14} & \numprint{0.0005} \\
		&\textbf{Modello 15} & \numprint{0.0005} \\	
		\bottomrule 
	\end{tabular}
	\captionof{table}{Learning rate delle simulazioni effettuate nell'esperimento 2}
	\label{tab:learningratemikolov2}
\end{table}

\subsection{Performance}
\label{subsec:performance4}

Prendendo in considerazione le diverse architetture implementate, vengono presentati i risultati ottenuti in termini di loss.

\begin{table}[H]
	\centering
	\begin{tabular}{ll@{\hspace{.5cm}}ccc}
		\toprule
		\multicolumn{2}{c}{{Modelli}} & \textbf{Train loss} & \textbf{Test loss} & \textbf{Tempo di training}  \\
		\midrule
		\multirow{3}*{{Embedding 2}} 
		&\textbf{Modello 11} & \numprint{3.275} & \numprint{3.455} &\numprint{20} h \\
		&\textbf{Modello 12} & \numprint{3.297} & \numprint{3.459} &\numprint{30} h \\
		&\textbf{Modello 13} & \numprint{3.444} & \numprint{3.458} &\numprint{40} h \\
		\midrule
		\multirow{2}*{{Embedding 2}} 
		&\textbf{Modello 14} & \numprint{3.350} & \numprint{3.459} &\numprint{20} h \\
		&\textbf{Modello 15} & \numprint{3.230} & \numprint{3.454} &\numprint{60} h \\	
		\bottomrule 
	\end{tabular}
	\captionof{table}{Confronto delle \emph{loss} ottenute nelle diverse reti con i due diversi embedding}
	\label{tab:lossmikolov3}
\end{table}

Come già detto, l'utilizzo di questo approccio consente una valutazione non più solo in termini di loss ma se ne potrà analizzare anche l'accuratezza, introdotta nella Sezione~\ref{subsubsec:classificazione}. 

\begin{figure}[H]
	\centering
	\begin{tabular}{clP{1cm}P{1cm}P{1cm}P{1cm}P{1cm}}
		\toprule	
		& 		 			& \multicolumn{5}{c}{\textbf{Train/Test Accuracy [\%]} 		}							       \\
		\multicolumn{2}{c}{\multirow{-2}{*}{Modelli}}
		& O 				& C 			   & E 				  & A 				 & N 			   \\ 
		\midrule
		\multirow{3}*{{Embedding 2}} 
		& \textbf{Modello 11} & \numprint{61}/\numprint{61} & \numprint{60}/\numprint{59} & \numprint{63}/\numprint{60} & \numprint{58}/\numprint{56} & \numprint{57}/\numprint{54} \\
		
		& \textbf{Modello 12} &\numprint{62}/\numprint{59} &\numprint{61}/\numprint{50} & \numprint{64}/\numprint{45} & \numprint{61}/\numprint{56} & \numprint{61}/\numprint{56} \\
		
		& \textbf{Modello 13} & \numprint{63}/\numprint{55} &\numprint{63}/\numprint{60} & \numprint{63}/\numprint{61} & \numprint{63}/\numprint{58} & \numprint{62}/\numprint{59} \\
		\midrule
		\multirow{2}*{{Embedding 3}} 
		& \textbf{Modello 14} & \numprint{61}/\numprint{61} & \numprint{61}/\numprint{57} & \numprint{62}/\numprint{52} & \numprint{60}/\numprint{57} & \numprint{60}/\numprint{56} \\
		
		& \textbf{Modello 15} &\numprint{63}/\numprint{60} & \numprint{62}/\numprint{48} & \numprint{64}/\numprint{60} & \numprint{62}/\numprint{49} & \numprint{62}/\numprint{48} \\
		\bottomrule	
	\end{tabular}
	\captionof{table}{Confronto delle accuracy sul test set delle diverse architetture}
	\label{tab:accuracymikolov}
\end{figure}

Il ``Modello 13'' sembrerebbe il migliore tra quelli proposti poiché presenta valori di accuracy più alti e la differenza di prestazioni tra train e test set è una tra le più basse, dimostrando minore propensione all'overfitting.  \\

Viene presentata un analisi più dettagliata su questo modello, in particolare mostrando il grafico dell'accuratezza, su train e test set, e le matrici di confusione calcolate per ogni tratto di personalità.

Come si può vedere chiaramente dalle matrici di confusione, le caratteristiche \emph{Conscientiousness} e \emph{Extraversion} sono quelle che vengono predette nel modo peggiore.

\begin{figure}[H]
	\centering
	\subfloat[][\emph{Visualizzazione matrice di confusione}\label{subfig:confO}]{
		\begin{minipage}[c][0.7\width]{0.45\textwidth}
			\centering
			\includegraphics[width=.6\textwidth]{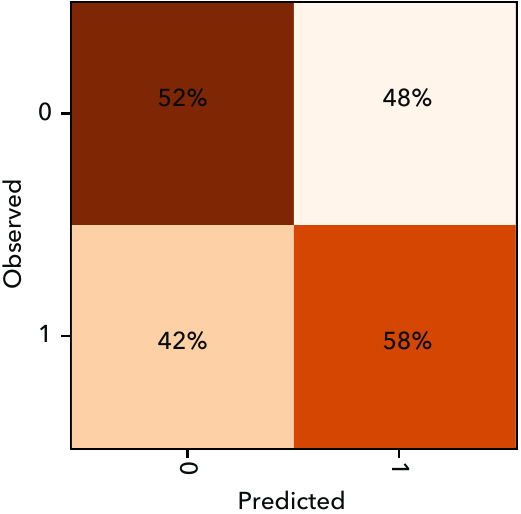}
		\end{minipage}} 
	\hspace{10mm}
	\subfloat[][\emph{Visualizzazione accuratezza su train e test}\label{subfig:accO}]{
		\begin{minipage}[c][0.7\width]{0.45\textwidth}
			\centering
			\includegraphics[width=.8\textwidth]{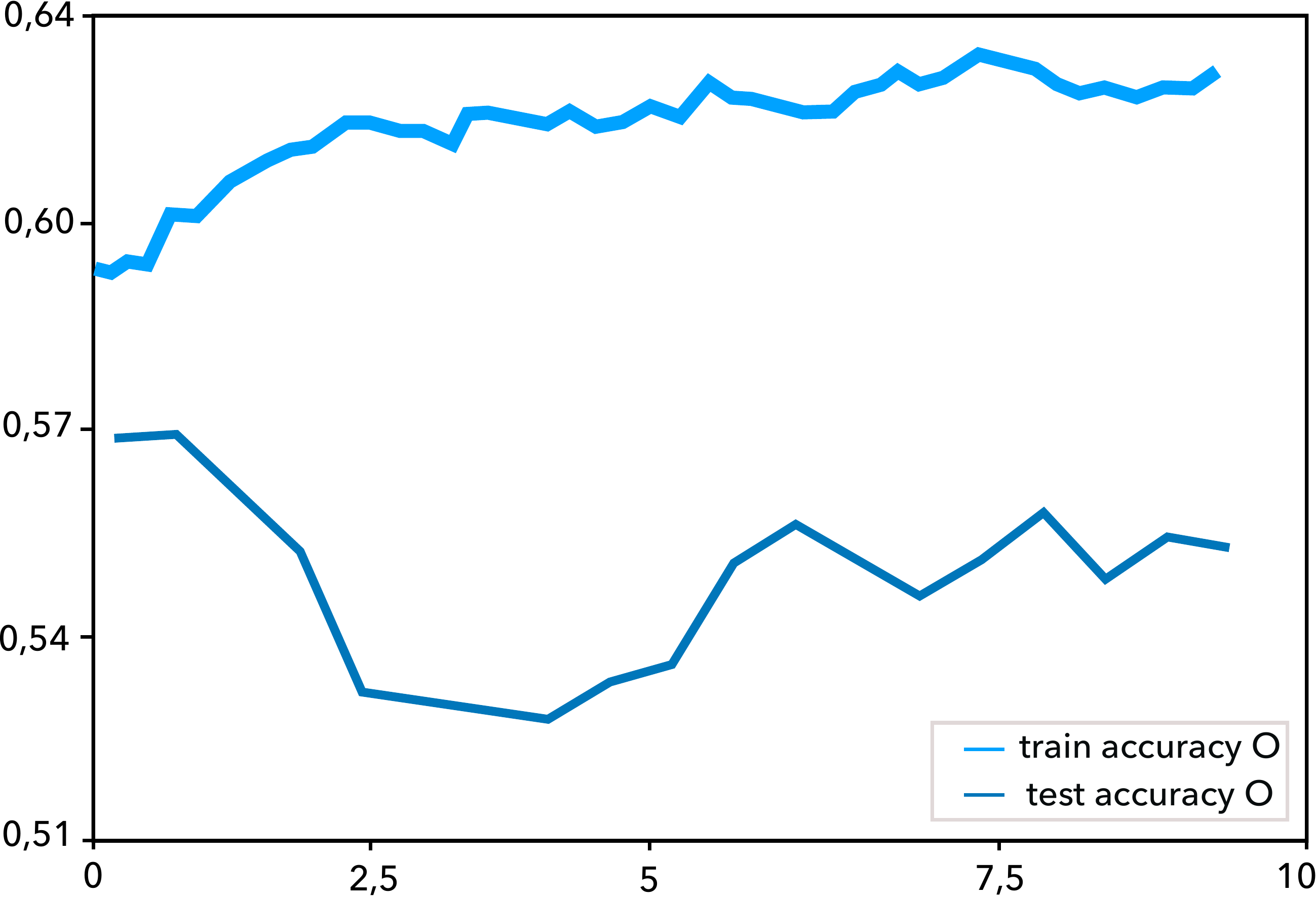}
		\end{minipage}}
	\caption{Analisi della caratteristica Openness del ``Modello 13''}
	\label{fig:binO}
\end{figure}

\begin{figure}[H]
	\centering
	\subfloat[][\emph{Visualizzazione matrice di confusione}\label{subfig:confC}]
	{\begin{minipage}[c][0.7\width]{0.45\textwidth}
			\centering
			\includegraphics[width=.6\textwidth]{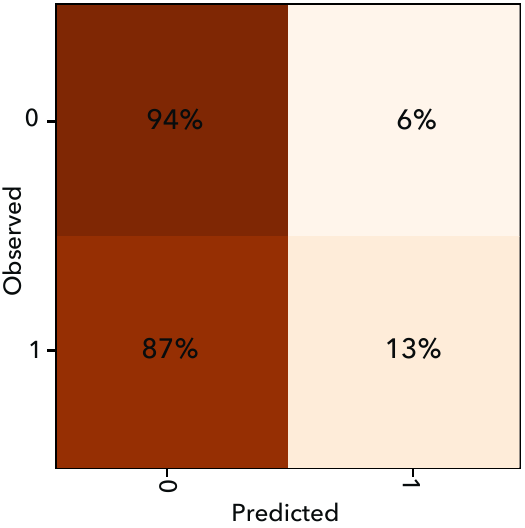}
		\end{minipage}} 
	\hspace{10mm}
	\subfloat[][\emph{Visualizzazione accuratezza su train e test}\label{subfig:accC}]
	{\begin{minipage}[c][0.7\width]{0.45\textwidth}
			\centering
			\includegraphics[width=.8\textwidth]{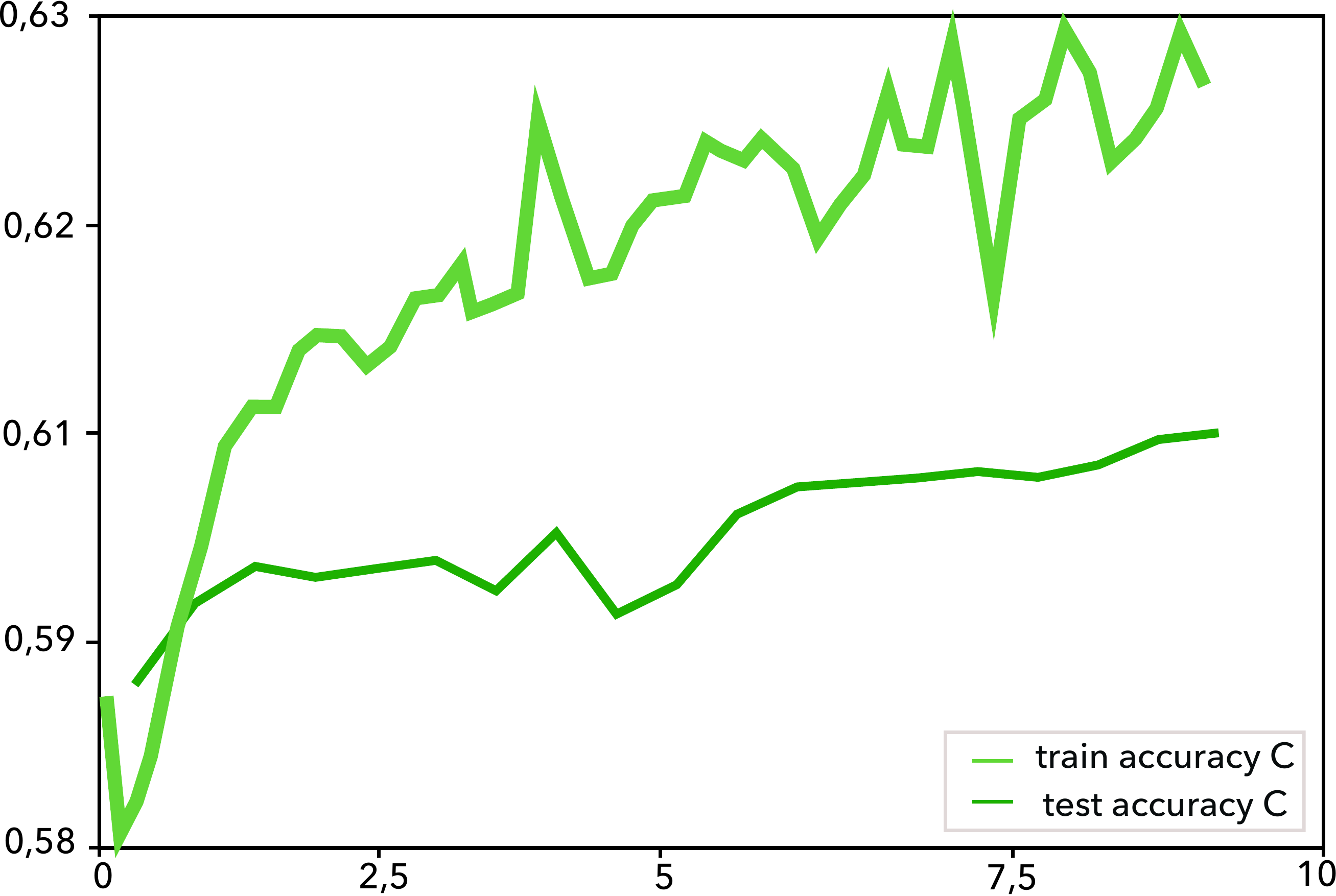}
		\end{minipage}}
	\caption{Analisi della caratteristica Conscientiousness del ``Modello 13''}
	\label{fig:binC}
\end{figure}

\begin{figure}[H]
	\centering
	\subfloat[][\emph{Visualizzazione matrice di confusione}\label{subfig:confE}]
	{\begin{minipage}[c][0.7\width]{0.45\textwidth}
			\centering
			\includegraphics[width=.6\textwidth]{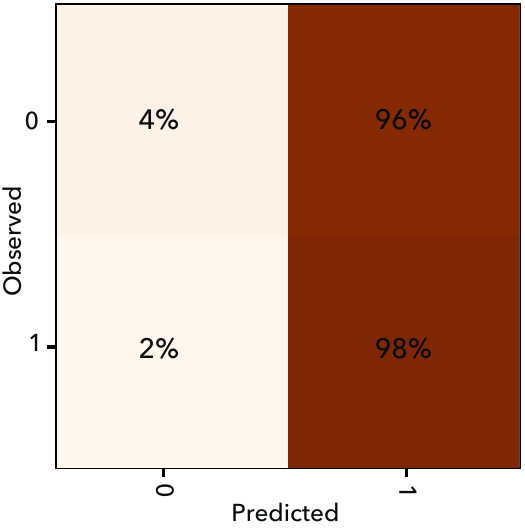}
		\end{minipage}} 
	\hspace{10mm}
	\subfloat[][\emph{Visualizzazione accuratezza su train e test}\label{subfig:accE}]
	{\begin{minipage}[c][0.7\width]{0.45\textwidth}
			\centering
			\includegraphics[width=.8\textwidth]{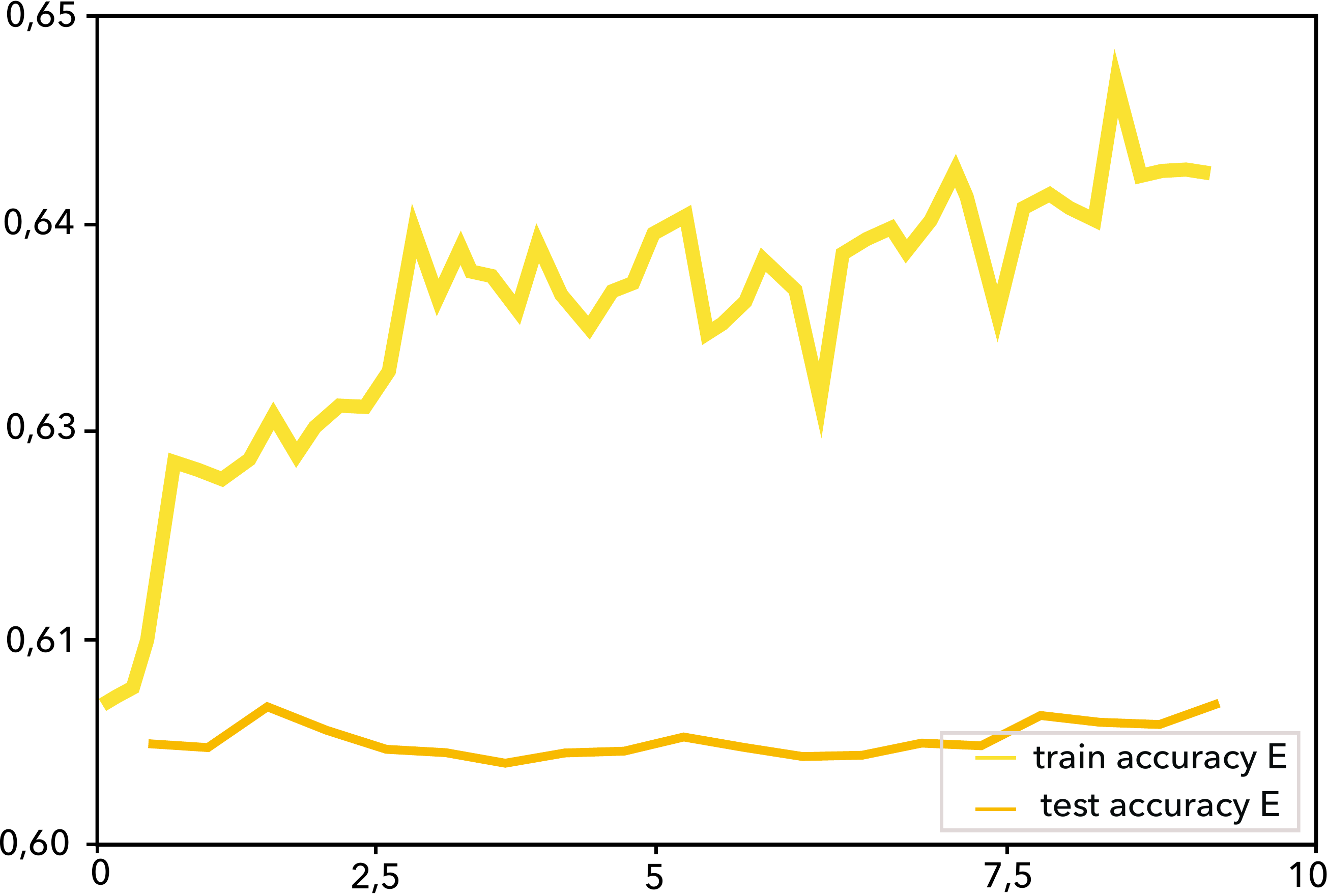}
		\end{minipage}}
	\caption{Analisi della caratteristica Extraversion del ``Modello 13''}
	\label{fig:binE}
\end{figure}

\begin{figure}[H]
	\centering
	\subfloat[][\emph{Visualizzazione matrice di confusione}\label{subfig:confA}]
	{\begin{minipage}[c][0.7\width]{0.45\textwidth}
			\centering
			\includegraphics[width=.6\textwidth]{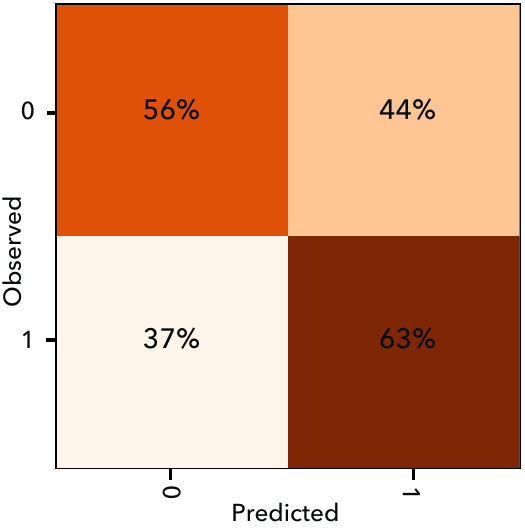}
		\end{minipage}} 
	\hspace{10mm}
	\subfloat[][\emph{Visualizzazione accuratezza su train e test}\label{subfig:accA}]
	{\begin{minipage}[c][0.7\width]{0.45\textwidth}
			\centering
			\includegraphics[width=.8\textwidth]{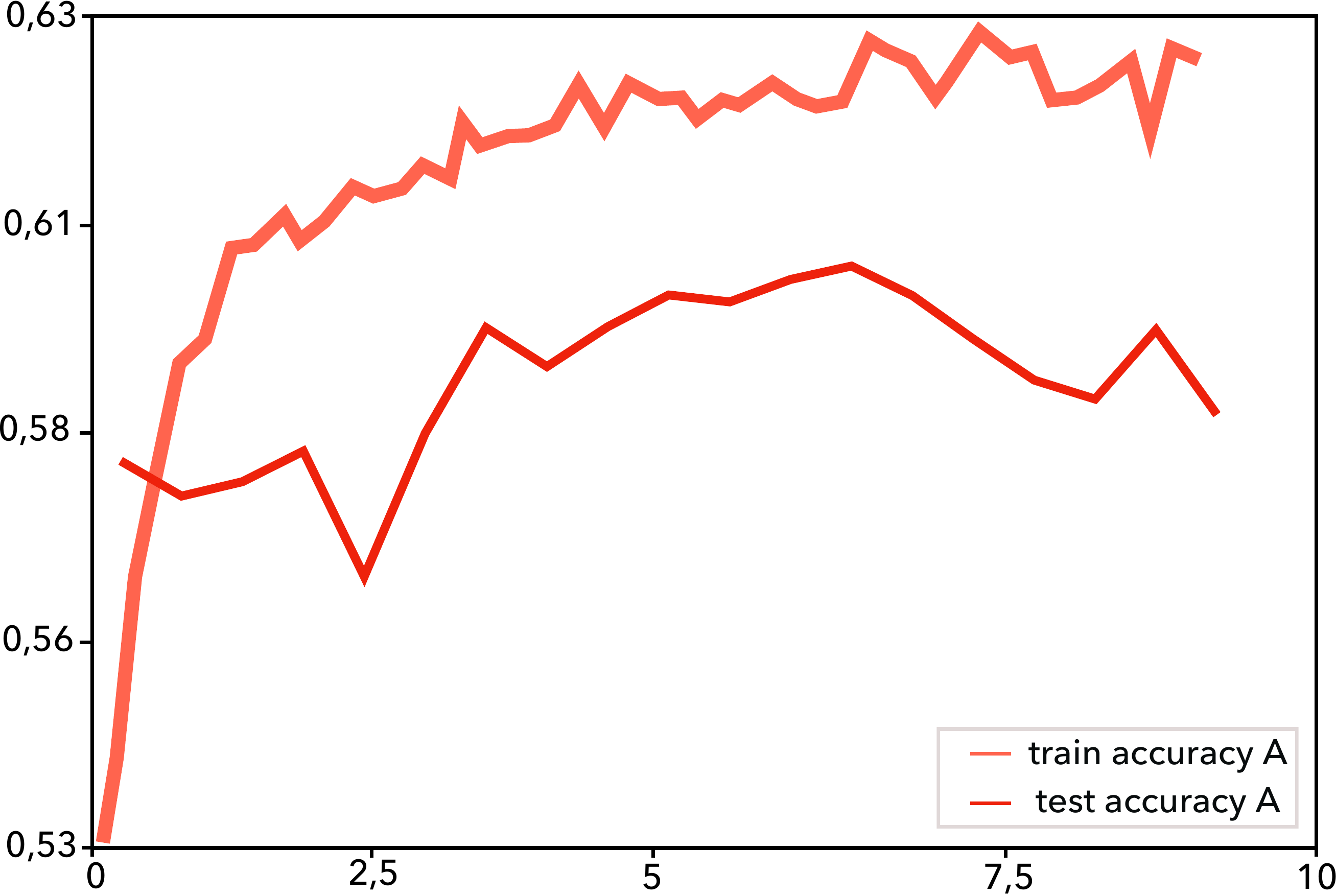}
		\end{minipage}}
	\caption{Analisi della caratteristica Agreeableness del ``Modello 13''}
	\label{fig:binA}
\end{figure}

\begin{figure}[H]
	\centering
	\subfloat[][\emph{Visualizzazione matrice di confusione}\label{subfig:confN}]
	{\begin{minipage}[c][0.7\width]{0.45\textwidth}
			\centering
			\includegraphics[width=.6\textwidth]{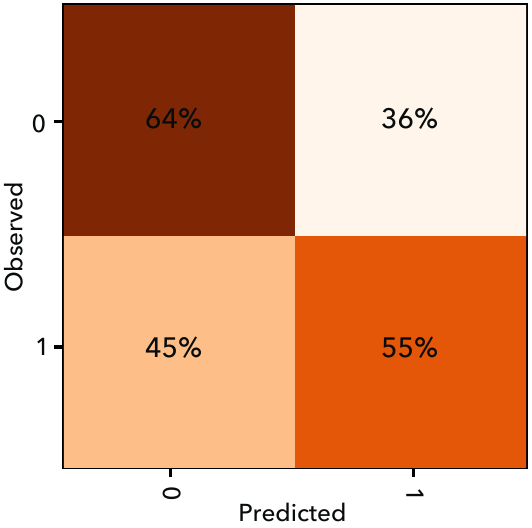}
		\end{minipage}} 
	\hspace{10mm}
	\subfloat[][\emph{Visualizzazione accuratezza su train e test}\label{subfig:accN}]
	{\begin{minipage}[c][0.7\width]{0.45\textwidth}
			\centering
			\includegraphics[width=.8\textwidth]{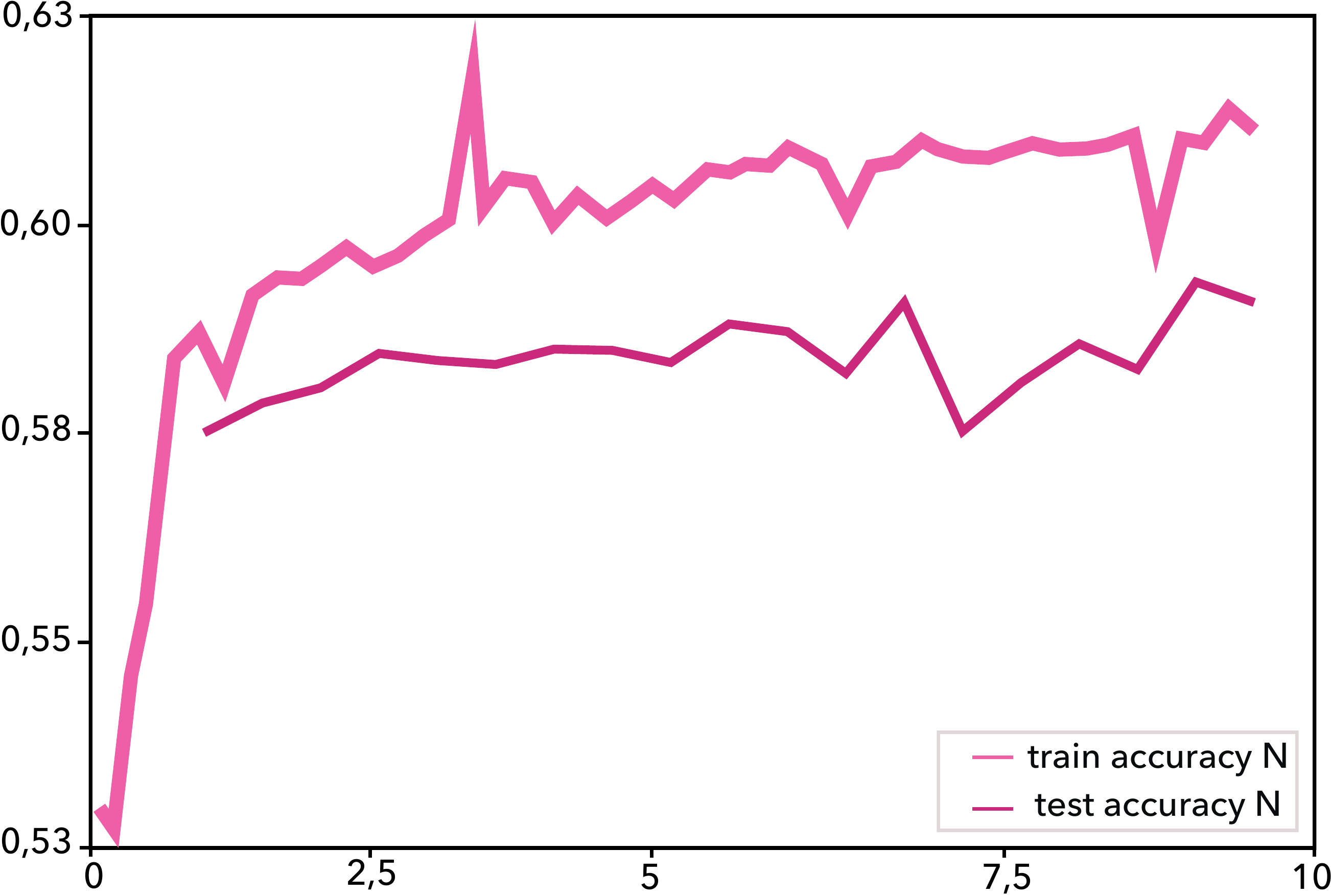}
		\end{minipage}}
	\caption{Analisi della caratteristica Neuroticism del ``Modello 13''}
	\label{fig:binN}
\end{figure}

\chapter{Conclusioni}
\label{chap:conclusioni}

La natura di questo progetto di tesi è altamente sperimentale ed è volta a presentare analisi dettagliate sull'argomento, in quanto allo stato attuale non esistono importanti indagini che affrontino il problema dell'apprendimento dei tratti di personalità a partire da testo in linguaggio naturale.
\\

La domanda fondamentale che viene posta al fine di risolvere questo compito è quale sia la metodologia adatta alla rappresentazione del testo.
Durante la sperimentazione, infatti, sono stati esaminati due principali approcci e ne è stata comparata l'efficacia.

Siamo partiti da una rappresentazione semplificata del testo ricorrendo al modello \emph{bag-of-words}. 
I risultati ottenuti sono sub-ottimali dal punto di vista della complessità. Inoltre la codifica utilizzata non fornisce alcuna informazione utile al sistema riguardo le relazioni che possono sussistere tra le parole di una frase.

In seguito è stata sfruttata una classe di algoritmi distribuzionali per insegnare alla rete il significato e le relazioni sussistenti tra le parole. Sfruttando la versione \emph{skip-gram} dell'algoritmo \texttt{word2vec} di Tomas Mikolov vengono rappresentate sotto forma di vettori le mappature tra parole e contesti nello spazio. 
Ricorrendo al secondo metodo i risultati ottenuti dimostrano come utilizzare un embedding sia la tecnica di estrazione di features più efficiente per filtrare le informazioni contenute in un testo.\\

È emerso che un'evoluzione di questo progetto potrebbe affacciarsi alla valutazione di rappresentazioni alternative del testo, annotazioni, part-of-speech \cite{brown1957linguistic} e altre tecniche di NLP, per approfondire e migliorare ulteriormente l'indagine.\\

Dal punto di vista delle architetture implementate, abbiamo iniziato dall'implementazione di reti neurali \emph{fully-connected} come base per capire come modelli semplici di Deep Learning possano fornire informazioni sulle caratteristiche nascoste della personalità. 

Infine, ci addentriamo nelle reti neurali \emph{convoluzionali} molto più specializzate e efficiente delle precedenti nell'ambito del Text Mining.

Sviluppi futuri di questo lavoro potrebbero valutare una procedura di apprendimento alternativa, sfruttando altri modelli, quali le reti ricorrenti,  per ottenere dei risultati più efficaci.\\

La prima valutazione effettuata è definita come una regressione che tenta di prevedere l'esatto valore reale per ogni tratto di personalità.
Il problema presentato è estremamente complesso, e le performance ottenute sono ancora lontane da quelle desiderate.
Per questo motivo viene eseguita una seconda valutazione che trasforma il nostro compito in un problema di classificazione binaria multi-label.
I risultati raggiunti questa volta, evincono che ottimizzare la funzione di loss di un modello predittivo di regressione è molto più difficoltoso rispetto ad ottimizzare una funzione di costo stabile come la Softmax.
Questo è evidente poiché il modello di regressione tenta di produrre un esatto valore per ogni input, e i valori anomali predetti possono introdurre gradienti enormi.



\printbibliography

@article{samuel1959some,
	title={Some studies in machine learning using the game of checkers},
	author={Samuel, Arthur L},
	journal={IBM Journal of research and development},
	volume={3},
	number={3},
	pages={210--229},
	year={1959},
	publisher={IBM}
}

@article{barrick1991big,
  title={The big five personality dimensions and job performance: a meta-analysis},
  author={Barrick, Murray R and Mount, Michael K},
  journal={Personnel psychology},
  volume={44},
  number={1},
  pages={1--26},
  year={1991},
  publisher={Wiley Online Library}
}

@article{goldberg1993structure,
  title={The structure of phenotypic personality traits.},
  author={Goldberg, Lewis R},
  journal={American psychologist},
  volume={48},
  number={1},
  pages={26},
  year={1993},
  publisher={American Psychological Association}
}

@article{costa2008revised,
	title={The revised neo personality inventory (neo-pi-r)},
	author={Costa, Paul T and McCrae, Robert R},
	journal={The SAGE handbook of personality theory and assessment},
	volume={2},
	number={2},
	pages={179--198},
	year={2008},
	publisher={Sage Publications Inc. Thousand Oaks, CA, US}
}

@article{viswesvaran2000measurement,
	title={Measurement error in “Big Five Factors” personality assessment: Reliability generalization across studies and measures},
	author={Viswesvaran, Chockalingam and Ones, Deniz S},
	journal={Educational and Psychological Measurement},
	volume={60},
	number={2},
	pages={224--235},
	year={2000},
	publisher={Sage Publications Sage CA: Thousand Oaks, CA}
}

@article{saulsman2004five,
  title={The five-factor model and personality disorder empirical literature: A meta-analytic review},
  author={Saulsman, Lisa M and Page, Andrew C},
  journal={Clinical psychology review},
  volume={23},
  number={8},
  pages={1055--1085},
  year={2004},
  publisher={Elsevier}
}

@article{franklin2005elements,
	title={The elements of statistical learning: data mining, inference and prediction},
	author={Franklin, James},
	journal={The Mathematical Intelligencer},
	volume={27},
	number={2},
	pages={83--85},
	year={2005},
	publisher={Springer}
}

@article{chakrabarti2006data,
	title={Data mining curriculum: A proposal (Version 1.0)},
	author={Chakrabarti, Soumen and Ester, Martin and Fayyad, Usama and Gehrke, Johannes and Han, Jiawei and Morishita, Shinichi and Piatetsky-Shapiro, Gregory and Wang, Wei},
	journal={Intensive Working Group of ACM SIGKDD Curriculum Committee},
	volume={140},
	year={2006}
}

@inproceedings{tan1999text,
	title={Text mining: The state of the art and the challenges},
	author={Tan, Ah-Hwee and others},
	booktitle={Proceedings of the PAKDD 1999 Workshop on Knowledge Disocovery from Advanced Databases},
	volume={8},
	pages={65--70},
	year={1999},
	organization={sn}
}

@book{corr2009cambridge,
	title={The Cambridge handbook of personality psychology},
	author={Corr, Philip J and Matthews, Gerald},
	year={2009},
	publisher={Cambridge University Press Cambridge}
}

@book{sadock2000comprehensive,
	title={Comprehensive textbook of psychiatry},
	author={Sadock, Benjamin J and Sadock, Virginia A and Ruiz, Pedro},
	year={2000},
	publisher={lippincott Williams \& wilkins Philadelphia}
}

@article{triandis2002cultural,
author = {Triandis, Harry C. and Suh, Eunkook M.},
title = {Cultural Influences on Personality},
journal = {Annual Review of Psychology},
volume = {53},
number = {1},
pages = {133-160},
year = {2002},
doi = {10.1146/annurev.psych.53.100901.135200},
eprint = {https://doi.org/10.1146/annurev.psych.53.100901.135200}
}

@inproceedings{nair2010rectified,
	title={Rectified linear units improve restricted boltzmann machines},
	author={Nair, Vinod and Hinton, Geoffrey E},
	booktitle={Proceedings of the 27th International Conference on machine learning (ICML-10)},
	pages={807--814},
	year={2010}
}

@article{hahnloser2000digital,
	author = {Hahnloser, Richard and Sarpeshkar, Rahul and A. Mahowald, Misha 
	and Douglas, Rodney and Sebastian Seung, H},
	year = {2000},
	month = {07},
	pages = {947-51},
	title = {Digital selection and analogue amplification coexist in a 
	cortex-inspired silicon circuit},
	volume = {405},
	booktitle = {Nature}
}

@inproceedings{hahnloser2003permitted,
 author = {Hahnloser, Richard and Seung, H. Sebastian},
 booktitle = {Advances in Neural Information Processing Systems},
 editor = {T. Leen and T. Dietterich and V. Tresp},
 pages = {},
 publisher = {MIT Press},
 title = {Permitted and Forbidden Sets in Symmetric Threshold-Linear Networks},
 url = {https://proceedings.neurips.cc/paper/2000/file/c8cbd669cfb2f016574e9d147092b5bb-Paper.pdf},
 volume = {13},
 year = {2000}
}

@inproceedings{glorot2011deep,
	title={Deep sparse rectifier neural networks},
	author={Glorot, Xavier and Bordes, Antoine and Bengio, Yoshua},
	booktitle={Proceedings of the fourteenth international conference on artificial intelligence and statistics},
	pages={315--323},
	year={2011}
}

@article{tang2013deep,
	title={Deep learning using linear support vector machines},
	author={Tang, Yichuan},
	journal={arXiv preprint arXiv:1306.0239},
	year={2013}
}

@article{duchi2011adaptive,
	title={Adaptive subgradient methods for online learning and stochastic optimization},
	author={Duchi, John and Hazan, Elad and Singer, Yoram},
	journal={Journal of Machine Learning Research},
	volume={12},
	number={Jul},
	pages={2121--2159},
	year={2011}
}

@book{burnham2003model,
	title={Model selection and multimodel inference: a practical information-theoretic approach},
	author={Burnham, Kenneth P and Anderson, David R},
	year={2003},
	publisher={Springer Science \& Business Media}
}

@inproceedings{mikolov2013distributed,
	title={Distributed representations of words and phrases and their compositionality},
	author={Mikolov, Tomas and Sutskever, Ilya and Chen, Kai and Corrado, Greg S and Dean, Jeff},
	booktitle={Advances in neural information processing systems},
	pages={3111--3119},
	year={2013}
}

@article{mikolov2013efficient,
	title={Efficient estimation of word representations in vector space},
	author={Mikolov, Tomas and Chen, Kai and Corrado, Greg and Dean, Jeffrey},
	journal={arXiv preprint arXiv:1301.3781},
	year={2013}
}

@inproceedings{mikolov2013linguistic,
	title={Linguistic regularities in continuous space word representations},
	author={Mikolov, Tomas and Yih, Wen-tau and Zweig, Geoffrey},
	booktitle={Proceedings of the 2013 Conference of the North American Chapter of the Association for Computational Linguistics: Human Language Technologies},
	pages={746--751},
	year={2013}
}

@book{manning1999foundations,
	title={Foundations of statistical natural language processing},
	author={Manning, Christopher D and Manning, Christopher D and Sch{\"u}tze, Hinrich},
	year={1999},
	publisher={MIT press}
}

@article{kim2014convolutional,
	title={Convolutional neural networks for sentence classification},
	author={Kim, Yoon},
	journal={arXiv preprint arXiv:1408.5882},
	year={2014}
}

@article{ioffe2015batch,
	title={Batch normalization: Accelerating deep network training by reducing internal covariate shift},
	author={Ioffe, Sergey and Szegedy, Christian},
	journal={arXiv preprint arXiv:1502.03167},
	year={2015}
}

@article{maaten2008visualizing,
	title={Visualizing data using t-SNE},
	author={Maaten, Laurens van der and Hinton, Geoffrey},
	journal={Journal of machine learning research},
	volume={9},
	number={Nov},
	pages={2579--2605},
	year={2008}
}

@article{liu2016classification,
	title={Classification with noisy labels by importance reweighting},
	author={Liu, Tongliang and Tao, Dacheng},
	journal={IEEE Transactions on pattern analysis and machine intelligence},
	volume={38},
	number={3},
	pages={447--461},
	year={2016},
	publisher={IEEE}
}

@article{wang2009mean,
	title={Mean squared error: Love it or leave it? A new look at signal fidelity measures},
	author={Wang, Zhou and Bovik, Alan C},
	journal={IEEE signal processing magazine},
	volume={26},
	number={1},
	pages={98--117},
	year={2009},
	publisher={IEEE}
}

@article{dyer2014notes,
	title={Notes on noise contrastive estimation and negative sampling},
	author={Dyer, Chris},
	journal={arXiv preprint arXiv:1410.8251},
	year={2014}
}

@inproceedings{liu2016large,
	title={Large-Margin Softmax Loss for Convolutional Neural Networks.},
	author={Liu, Weiyang and Wen, Yandong and Yu, Zhiding and Yang, Meng},
	booktitle={ICML},
	pages={507--516},
	year={2016}
}

@article{ruder2016overview,
	title={An overview of gradient descent optimization algorithms},
	author={Ruder, Sebastian},
	journal={arXiv preprint arXiv:1609.04747},
	year={2016}
}

@article{horikawa1992fuzzy,
	title={On fuzzy modeling using fuzzy neural networks with the back-propagation algorithm},
	author={Horikawa, S-I and Furuhashi, Takeshi and Uchikawa, Yoshiki},
	journal={IEEE transactions on Neural Networks},
	volume={3},
	number={5},
	pages={801--806},
	year={1992},
	publisher={IEEE}
}

@inproceedings{wallach2006topic,
	title={Topic modeling: beyond bag-of-words},
	author={Wallach, Hanna M},
	booktitle={Proceedings of the 23rd international conference on Machine learning},
	pages={977--984},
	year={2006},
	organization={ACM}
}

@article{svozil1997introduction,
	title={Introduction to multi-layer feed-forward neural networks},
	author={Svozil, Daniel and Kvasnicka, Vladimir and Pospichal, Jiri},
	journal={Chemometrics and intelligent laboratory systems},
	volume={39},
	number={1},
	pages={43--62},
	year={1997},
	publisher={Elsevier}
}

@inproceedings{sainath2015convolutional,
	title={Convolutional, long short-term memory, fully connected deep neural networks},
	author={Sainath, Tara N and Vinyals, Oriol and Senior, Andrew and Sak, Ha{\c{s}}im},
	booktitle={Acoustics, Speech and Signal Processing (ICASSP), 2015 IEEE International Conference on},
	pages={4580--4584},
	year={2015},
	organization={IEEE}
}

@article{karpathy2016cs231n,
	title={Cs231n convolutional neural networks for visual recognition},
	author={Karpathy, Andrej},
	journal={Neural networks},
	volume={1},
	year={2016}
}

@inproceedings{erk2008structured,
	title={A structured vector space model for word meaning in context},
	author={Erk, Katrin and Pad{\'o}, Sebastian},
	booktitle={Proceedings of the Conference on Empirical Methods in Natural Language Processing},
	pages={897--906},
	year={2008},
	organization={Association for Computational Linguistics}
}

@inproceedings{baroni2014don,
	title={Don't count, predict! A systematic comparison of context-counting vs. context-predicting semantic vectors},
	author={Baroni, Marco and Dinu, Georgiana and Kruszewski, Germ{\'a}n},
	booktitle={Proceedings of the 52nd Annual Meeting of the Association for Computational Linguistics (Volume 1: Long Papers)},
	volume={1},
	pages={238--247},
	year={2014}
}

@article{brown1957linguistic,
	title={Linguistic determinism and the part of speech.},
	author={Brown, Roger W},
	journal={The Journal of Abnormal and Social Psychology},
	volume={55},
	number={1},
	pages={1},
	year={1957},
	publisher={American Psychological Association}
}
\addcontentsline{toc}{chapter}{\refname}

\end{document}